\def\enactoptarxiv{}
\documentclass{article} % For LaTeX2e
\usepackage{iclr2027_conference,times}

\usepackage{amsmath,amsfonts,bm}

\def\eqref#1{equation~\ref{#1}}
\def\1{\bm{1}}

\DeclareMathAlphabet{\mathsfit}{\encodingdefault}{\sfdefault}{m}{sl}
\SetMathAlphabet{\mathsfit}{bold}{\encodingdefault}{\sfdefault}{bx}{n}

\usepackage{hyperref}
\usepackage{url}
\usepackage{graphicx}
\usepackage{booktabs}
\usepackage{tabularx}
\usepackage{colortbl}
\usepackage{etoc}
\usepackage{textcomp}
\usepackage[T1,OT1]{fontenc}
\usepackage{tcolorbox}
\tcbuselibrary{listings,skins,breakable}

\definecolor{enactoptrow}{RGB}{234,242,248}
\definecolor{mainheaderrow}{RGB}{223,234,250}
\definecolor{mainenactoptrow}{RGB}{249,221,228}
\definecolor{mainenactoptgain}{RGB}{0,108,57}
\definecolor{tablegain}{RGB}{0,120,65}
\definecolor{tableloss}{RGB}{180,40,40}
\definecolor{harnessboxframe}{RGB}{47,79,111}
\definecolor{harnessboxfill}{RGB}{247,249,252}

\newtcblisting{harnesslisting}[2][9.5]{%
	enhanced jigsaw,
	breakable,
	listing only,
	colback=harnessboxfill,
	colframe=harnessboxframe,
	colbacktitle=harnessboxframe,
	coltitle=white,
	fonttitle=\small\bfseries,
	title={#2},
	title after break={#2\space(continued)},
	arc=1.2mm,
	boxrule=0.5pt,
	boxsep=0pt,
	left=3pt,
	right=3pt,
	lefttitle=8pt,
	righttitle=8pt,
	top=6pt,
	bottom=6pt,
	toptitle=3pt,
	bottomtitle=3pt,
	before skip=10pt,
	after skip=12pt,
	listing options={%
		basicstyle=\fontsize{8}{#1}\fontencoding{T1}\ttfamily\selectfont,
		columns=fullflexible,
		keepspaces=true,
		breaklines=true,
		breakatwhitespace=true,
		breakindent=1em,
		showstringspaces=false,
		tabsize=4,
		upquote=true,
		aboveskip=0pt,
		belowskip=0pt,
		literate={—}{{\textemdash}}1
		         {’}{{\textquoteright}}1
		         {…}{{\textellipsis}}1
	}
}

\newcommand{\mainresult}[2]{%
	#1\makebox[0pt][l]{\hspace{0.12em}\raisebox{-0.35ex}{\small\textcolor{tablegain}{#2}}}%
}

\title{EnactOpt: Diagnosis-Guided\\Harness Evolution for Embodied Agents}

\author{%
ZhiYuan Feng\textsuperscript{1}\thanks{Equal contribution.}\enspace
\thanks{Work done during an internship at Kling AI.}\quad
Zhuoshi Pan\textsuperscript{1}\footnotemark[1]\enspace\footnotemark[2]\quad
Zhaolu Kang\textsuperscript{2}\footnotemark[1]\quad
Zhongxia Chen\textsuperscript{3}\thanks{Project leads.}\\
\bfseries Shuo Yang\textsuperscript{3}\footnotemark[3]\quad
Sen Ye\textsuperscript{2}\quad
Heng Wang\textsuperscript{4}\quad
Han Li\textsuperscript{3}\\[0.5ex]
{\normalfont\small\textsuperscript{1}Tsinghua University\quad
\textsuperscript{2}Peking University}\\
{\normalfont\small\textsuperscript{3}Kling AI\quad
\textsuperscript{4}University of Illinois at Urbana-Champaign}
}

\ifdefined\enactoptarxiv
\iclrfinalcopy
\definecolor{citationblue}{RGB}{43,107,201}
\hypersetup{
	colorlinks=true,
	citecolor=citationblue,
	linkcolor=black,
	urlcolor=black,
	pdftitle={EnactOpt: Diagnosis-Guided Harness Evolution for Embodied Agents},
	pdfauthor={ZhiYuan Feng, Zhuoshi Pan, Zhaolu Kang, Zhongxia Chen, Shuo Yang, Sen Ye, Heng Wang, Han Li}
}
\fi

\newcommand{\appendixref}[1]{%
	\ifdefined\enactoptarxiv
		\begingroup
		\hypersetup{linkcolor=citationblue}%
		\hyperref[#1]{Appendix~\ref*{#1}}%
		\endgroup
	\else
		Appendix~\ref{#1}%
	\fi
}
\begin{document}

\maketitle
\ifdefined\enactoptarxiv
% Override the style's publication header for the unreviewed preprint.
% 覆盖模板的已发表页眉，使预印本不带会议发表声明。
\pagestyle{plain}
\fi

\begin{abstract}
Embodied agents built on foundation models rely on an execution harness
to organize observations, guide actions, and incorporate execution
feedback into subsequent decisions. Failures can arise not only from
limited model capabilities, but also from how the harness presents
evidence and guides its use. These model-external mechanisms provide
an adaptation target without updating model parameters, but task-specific
interaction experience can be costly to collect. Task-level outcomes
alone do not reveal whether relevant evidence was missing from the
model input or was available but inadequately used. Moreover, these
two aspects are coupled: effective action guidance may require evidence
that the current context omits, while exposing that evidence alone does
not ensure an appropriate response. We introduce EnactOpt, which uses
limited execution feedback to jointly evolve reusable textual skills
and context-construction code while keeping the underlying models and
runtime fixed. It diagnoses adjacent decisions by relating action
feedback to the actual model input and subsequent action. Structured
episode summaries guide joint revisions through prompt-level edit
operators, and candidate harnesses undergo compatibility checks and
validation-based retention. Across VLABench, ALFWorld, and Embodied
Agent Interface (EAI), EnactOpt improves task success by approximately
10--27 percentage points over the initial harnesses and
outperforms the evaluated GEPA, TextGrad, and Trace2Skill baselines.
These gains extend to task conditions not used for harness evolution
or candidate selection. The results support a complementary route to
embodied adaptation: turning limited execution experience into reusable
behavioral improvements by jointly refining how evidence is presented
and acted upon.
% 中文对照：
% 基于基础模型的具身智能体依赖执行支架组织观测、指导行动，并将执行反馈纳入后续决策。
% 失败不仅可能来自模型能力有限，也可能来自支架呈现证据和指导证据使用的方式。
% 这些模型外机制提供了无需更新模型参数的适配对象，但获取任务相关交互经验可能代价高昂。
% 仅凭任务级结果，无法判断相关证据是没有进入模型输入，还是已经提供却未被恰当利用。
% 此外，证据呈现与行动指导相互耦合：有效的行动指导可能需要当前上下文遗漏的证据，
% 而仅仅呈现这些证据，也不能保证智能体作出恰当响应。
% 我们提出 EnactOpt，利用有限执行反馈联合演化可复用文本技能与上下文构造代码，
% 同时保持底层模型与运行框架固定。
% 它将动作反馈与实际模型输入及后续动作关联起来，对相邻决策进行诊断。
% 结构化 episode 总结通过提示词层面的编辑算子指导联合修订，
% 候选 harness 经过兼容性检查，并依据验证表现保留。
% 在 VLABench、ALFWorld 和 Embodied Agent Interface（EAI）上，EnactOpt 的任务成功率
% 比初始 harness 高约 10--27 个百分点，并超过本研究评估的 GEPA、TextGrad 和 Trace2Skill 基线。
% 这些收益延伸至未用于 harness 演化或候选选择的任务条件。
% 结果支持一条互补的具身适配路径：通过协同改进证据的呈现与使用方式，
% 将有限执行经验转化为可复用的行为改进。
\end{abstract}

\etocdepthtag.toc{main}
\section{Introduction}
\label{sec:introduction}
\begin{figure}[t]
\centering
\includegraphics[width=\linewidth]{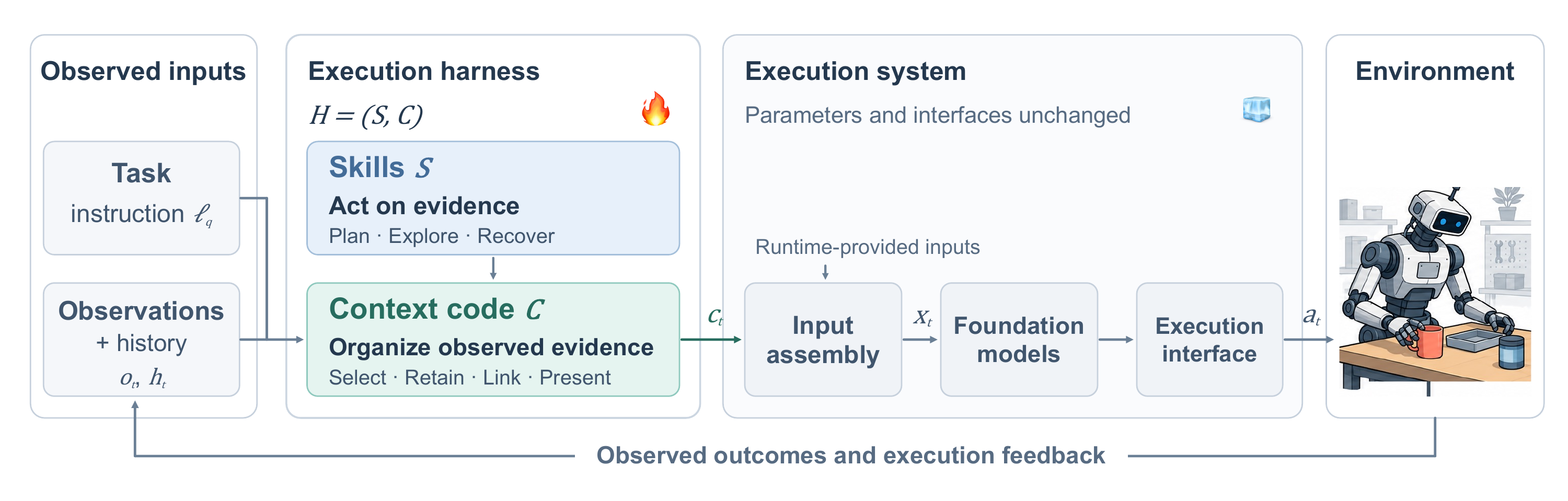}
\caption{Execution harness for embodied agents.
The editable harness $H=(S,C)$ comprises textual skills $S$ for action
guidance and context-construction code $C$ for organizing observed evidence.
The fixed runtime assembles the complete model input $x_t$ from constructed
context $c_t$ and any additional runtime-provided inputs; model parameters
and runtime interfaces remain unchanged during harness evolution.}
% 中文对照：
% 具身智能体的执行支架。
% 可编辑的 harness $H=(S,C)$ 包含指导行动的文本技能 $S$，
% 以及组织已观察证据的上下文构造代码 $C$。
% 固定运行框架将构造的上下文 $c_t$ 与可能额外提供的输入组装为完整模型输入 $x_t$；
% 模型参数与运行接口在 harness 演化期间保持不变。
\label{fig:embodied-agent}
\end{figure}

Foundation models provide embodied agents with multimodal reasoning
\citep{driess2023palm,hurst2024gpt,bai2025qwen3,comanici2025gemini}
and sensorimotor control
\citep{kim2024openvla,intelligence2025pi_,li2025scalable,ye2026world}.
Effective execution, however, depends on how observations become model
inputs, model outputs become actions, and execution feedback informs
subsequent decisions \citep{shi2025hi}.
We call the reusable, model-external guidance and control logic supporting
this interaction the execution harness.
We focus on two editable components (Figure~\ref{fig:embodied-agent}):
reusable textual skills $S$ that guide actions and context-construction
code $C$ that selects and presents observed evidence.
Model parameters and the execution interface remain fixed during
harness adaptation.
% 中文对照：
% 基础模型为具身智能体提供多模态推理能力
% \citep{driess2023palm,hurst2024gpt,bai2025qwen3,comanici2025gemini}，
% 以及感知运动控制能力 \citep{kim2024openvla,intelligence2025pi_,li2025scalable,ye2026world}。
% 然而，有效执行还取决于观测如何成为模型输入、模型输出如何转化为动作，
% 以及执行反馈如何影响后续决策 \citep{shi2025hi}。
% 我们将支持这一交互过程的、位于模型外部的可复用指导与控制逻辑称为执行支架。
% 本文关注其中两个可编辑组件（图 \ref{fig:embodied-agent}）：
% 指导行动的可复用文本技能 $S$，以及选择并呈现已观察证据的上下文构造代码 $C$。
% 在 harness 适配期间，模型参数与执行接口保持固定。

An initial harness encodes assumptions about how tasks are expressed
and what evidence is available during execution. Changes in instructions
or visual conditions can make its action guidance or evidence selection
inadequate. Revising the harness still requires task-specific experience,
which can be costly to collect in physical environments
\citep{kim2024openvla,tang2025deep}.
Prior work on reflection and experience-guided learning shows that such
experience can be retained as reusable guidance
\citep{shinn2023reflexion,zhao2024expel}.
The challenge is therefore to turn limited execution evidence into
revisions that improve the harness on later tasks, rather than corrections
confined to the episode that produced it.
% 中文对照：
% 初始 harness 隐含了对任务表达方式及执行中可用证据的假设。
% 指令或视觉条件的变化可能使其行动指导或证据选择不再合适。
% 修订 harness 仍然需要任务相关经验，而在物理环境中收集这类经验可能代价高昂
% \citep{kim2024openvla,tang2025deep}。
% 已有反思与经验引导学习工作表明，这类经验可以被保留为可复用的指导
% \citep{shinn2023reflexion,zhao2024expel}。
% 因此，挑战在于将有限执行证据转化为能够改善后续任务执行的 harness 修订，
% 而不是仅针对产生这些经验的当前 episode 进行纠错。

Feedback-driven adaptation already spans reusable skill optimization
\citep{yang2026skillopt,ju2026embodiskill} and executable harness evolution
\citep{lee2026metaharness,lin2026agentic}, including skill-and-context
adaptation for embodied agents \citep{wang2026shaper}.
Useful revisions, however, require more than a task-level success signal.
A failure may reflect feedback omitted from the actual model input,
an inadequate response to available feedback, or both. This motivates
examining what evidence reached each decision and how the agent responded.
At the same time, a proposed skill revision may depend on observed
evidence not yet presented to the model. Evaluating that revision together
with the corresponding change to context construction allows their
combined effect to be assessed as one harness candidate, rather than
treating the two edits as independent improvements.
% 中文对照：
% 反馈驱动的适应已覆盖可复用技能优化
% \citep{yang2026skillopt,ju2026embodiskill} 和可执行 harness 演化
% \citep{lee2026metaharness,lin2026agentic}，
% 也包括具身智能体中的技能与上下文适应 \citep{wang2026shaper}。
% 然而，有用的修订需要的不只是任务级成功信号。
% 失败可能反映反馈未进入实际模型输入、对已有反馈响应不当，或两者同时存在。
% 因此，需要检查每次决策实际获得了哪些证据，以及智能体如何响应。
% 同时，拟议的技能修订可能依赖已经观察到、但尚未呈现给模型的证据。
% 将这项修订与对应的上下文构造修改组成同一个 harness 候选进行评估，
% 可以直接考察二者的组合效果，而非将两项编辑视为彼此独立的改进。

We introduce EnactOpt, a framework for diagnosis-guided joint revision
of textual skills and context-construction code (Figure~\ref{fig:pipeline}).
It diagnoses adjacent decisions in training rollouts, relating action
feedback to the actual model input and the agent's response.
Structured episode summaries consolidate the reported evidence,
uncertainty, and repair and preservation suggestions into feedback
for a single LLM optimizer. Given the source harness and this feedback,
the optimizer uses prompt-guided edit operators to organize targeted
changes to $S$, $C$, or both within one harness candidate.
Interface and runtime compatibility checks screen candidates before
validation-based pool retention. The retained versions supply subsequent
iterations, and the final validation-selected harness is fixed for testing.
% 中文对照：
% 我们提出 EnactOpt，一个对文本技能与上下文构造代码进行诊断引导联合修订的框架，
% 如图 \ref{fig:pipeline} 所示。
% 它对训练轨迹中的相邻决策进行诊断，将动作反馈与实际模型输入及智能体响应关联起来。
% 结构化 episode 总结将所报告的证据、不确定性、修复与保留建议整合为单一 LLM 优化器的反馈。
% 优化器接收来源 harness 及其反馈，通过提示引导的编辑算子组织针对性修改，
% 在同一个 harness 候选中修订 $S$、$C$ 或两者。
% 候选先经过接口与运行兼容性检查，再依据验证表现保留。
% 保留版本供后续迭代使用，最终通过验证选出的 harness 固定用于测试。

We evaluate benchmark-specific EnactOpt harnesses on VLABench
\citep{zhang2025vlabench}, ALFWorld \citep{shridhar2020alfworld}, and
Embodied Agent Interface (EAI) \citep{li2024embodied}, spanning low-level
policy execution and high-level action interfaces.
Under the reported configurations, EnactOpt improves task success by
approximately 10--27 percentage points over the initial harnesses and
outperforms the evaluated GEPA \citep{agrawal2026gepa},
TextGrad \citep{yuksekgonul2024textgrad}, and
Trace2Skill \citep{ni2026trace2skill} baselines on all three benchmarks
(Table~\ref{tab:in-distribution-results}).
The gains extend to within-benchmark task conditions not used for
evolution or candidate selection. On VLABench, the ablation comparisons
show larger performance gaps under shifted conditions than
in-distribution. Joint revision also achieves higher generalization
success with fewer optimization passes than separate skill and code
updates (Table~\ref{tab:vlabench-ablation}).
% These comparisons remain exploratory: they do not establish gains under
% matched computation or isolate the contributions of individual design choices.
% 中文对照：
% 我们在 VLABench \citep{zhang2025vlabench}、ALFWorld \citep{shridhar2020alfworld} 和
% Embodied Agent Interface（EAI）\citep{li2024embodied} 上评估各基准分别适配的 EnactOpt harness，
% 覆盖低层策略执行和高层动作接口。
% 在所报告的配置下，EnactOpt 的任务成功率比初始 harness 高约 10--27 个百分点，
% 并在三个基准上均优于所评估的 GEPA \citep{agrawal2026gepa}、
% TextGrad \citep{yuksekgonul2024textgrad} 和
% Trace2Skill \citep{ni2026trace2skill} 基线（表 \ref{tab:in-distribution-results}）。
% 这些收益延伸至未用于演化或候选选择的基准内部任务条件。
% 在 VLABench 上，消融比较显示条件变化下的表现差距比分布内更大；
% 联合修订还以更少的优化循环取得了比技能与代码分开更新更高的泛化成功率
% （表 \ref{tab:vlabench-ablation}）。
% 这些比较仍是探索性的：它们不能确立等计算量下的收益，也不能单独识别各项设计的贡献。

Our contributions are threefold:
% 中文对照：
% 我们的贡献分为三点：
\begin{itemize}
\item We introduce EnactOpt, a framework that treats textual skills and
context-construction code as a coupled adaptation target. A single
proposal can revise either or both components, enabling their combined
behavior to be evaluated while models and runtime remain fixed.
% 中文对照：
% 我们提出 EnactOpt，将文本技能与上下文构造代码作为相互耦合的适配对象。
% 在模型与运行框架固定的条件下，单次提案可以修订任一组件或两者，
% 从而评估配套修改后的整体行为。

\item We develop a diagnosis-to-revision feedback interface that relates
action feedback to actual model inputs and subsequent responses.
Structured episode summaries convey reported evidence and uncertainty,
while prompt-guided edit operators organize targeted changes and
preservation requests.
% 中文对照：
% 我们设计连接诊断与修订的反馈接口，将动作反馈与实际模型输入及后续响应关联起来。
% 结构化 episode 总结传递所报告的证据与不确定性，
% 提示引导的编辑算子则组织针对性修改与行为保留要求。

\item We report gains over initial harnesses and evaluated baselines
across three embodied benchmarks, including gains beyond the evolution
conditions. VLABench ablations show that comparable in-distribution
success can mask differences in generalization, and that joint revision
attains higher generalization success with fewer optimization passes
than separate updates.
% 中文对照：
% 我们在三个具身基准上报告相对初始 harness 与所评估基线的收益，
% 其中包括超出演化条件的收益。
% VLABench 消融表明，相近的分布内成功率可能掩盖泛化表现的差异；
% 联合修订则以更少的优化循环取得了比分开更新更高的泛化成功率。
\end{itemize}

\section{Related Work}
\label{sec:related-work}

\subsection{Embodied Agent Adaptation}
\label{sec:related-embodied-agents}
% 中文标题：具身智能体适应。
Adaptation in embodied agents spans adjustments during execution and
updates retained for later tasks. Hierarchical models and code-based
policies connect task-level instructions to executable behavior
\citep{shi2025hi,liang2023code}.
TaskGround grounds complete household scenes into task-relevant
context and infers executable task structure from situated requests
before generating skill-level action sequences
\citep{feng2026taskground}.
Closed-loop systems use observations and execution feedback to revise
plans and recover from failures \citep{huang2022inner,zhi2025closed}.
Beyond these within-task adjustments,
policy parameters can be adapted through demonstration-based fine-tuning
\citep{kim2024openvla} or reinforcement learning from deployment experience
\citep{intelligence2025pi06}. Experience can also be retained outside model
parameters: DROC distills human language corrections into retrievable
knowledge \citep{zha2024distilling}, whereas ASPIRE autonomously refines
robot control programs and consolidates validated repairs into a reusable
skill library \citep{lu2026aspire}. EnactOpt follows this latter direction
by jointly revising reusable textual skills and context-construction code
while keeping the underlying models fixed.
% 中文对照：
% 具身智能体的适配既包括执行过程中的调整，也包括为后续任务保留的更新。
% 分层模型与基于代码的策略将任务级指令连接到可执行行为
% \citep{shi2025hi,liang2023code}。
% TaskGround 将完整家庭场景组织为任务相关上下文，从情境化请求中推断可执行任务结构，
% 再生成技能级动作序列 \citep{feng2026taskground}。
% 闭环系统则利用观测与执行反馈修订计划并从失败中恢复
% \citep{huang2022inner,zhi2025closed}。
% 除了这些任务内调整，策略参数还可以通过基于示范的微调 \citep{kim2024openvla}，
% 或利用部署经验的强化学习 \citep{intelligence2025pi06} 进行适配。
% 经验也可以保留在模型参数之外：DROC 将人的语言纠正提炼为可检索知识
% \citep{zha2024distilling}，ASPIRE 则自主修订机器人控制程序，
% 并将经过验证的修复整合到可复用技能库中 \citep{lu2026aspire}。
% EnactOpt 沿着后一条路线，在保持底层模型固定的情况下，
% 对可复用文本技能与上下文构造代码进行联合修订。

\subsection{Agent Harness Optimization}
\label{sec:related-agent-evolution}
\label{sec:related-experience-skills}
% 中文标题：智能体支架优化。
Feedback-driven optimization differs in both the artifact revised and
how experience is consolidated. ExpeL and Voyager retain experience as
reusable natural-language guidance and executable skills, respectively
\citep{zhao2024expel,wang2023voyager}. TextGrad propagates textual feedback
through computation graphs, while GEPA evolves prompts through
trajectory-based reflection \citep{yuksekgonul2024textgrad,agrawal2026gepa}.
For reusable guidance, ACE incrementally updates structured playbooks,
whereas Trace2Skill consolidates lessons from a pool of trajectories into
skills \citep{zhang2026ace,ni2026trace2skill}. Skill revision further
addresses update control and attribution through bounded, validation-gated
edits \citep{yang2026skillopt}, distinctions between skill defects and
execution lapses \citep{ju2026embodiskill}, and preservation-aware
structural reorganization informed by accumulated diagnoses and
success--failure contrasts \citep{wang2026skillgrad}.
% 中文对照：
% 反馈驱动的优化在修订对象和经验整合方式上有所不同。
% ExpeL 与 Voyager 分别将经验保留为可复用的自然语言指导和可执行技能
% \citep{zhao2024expel,wang2023voyager}。
% TextGrad 在计算图中传播文本反馈，GEPA 则通过基于轨迹的反思演化提示
% \citep{yuksekgonul2024textgrad,agrawal2026gepa}。
% 对于可复用指导，ACE 增量更新结构化策略手册，Trace2Skill 则将轨迹池中的经验整合为技能
% \citep{zhang2026ace,ni2026trace2skill}。
% 技能修订进一步从更新控制与归因两个方面改进修订过程：
% 使用有界且经验证筛选的编辑 \citep{yang2026skillopt}，
% 区分技能缺陷与执行失误 \citep{ju2026embodiskill}，
% 或依据累积诊断和成功/失败对照，进行考虑有效行为保留的结构重组 \citep{wang2026skillgrad}。

For executable harnesses, AFlow formulates workflow construction as
search over code \citep{zhang2025aflow}. A central distinction is the
evidence exposed to the optimizer: Meta-Harness provides access to prior
code, scores, and raw trajectories, whereas AHE organizes experience into
a layered evidence corpus and links edits to predicted effects through
change manifests \citep{lee2026metaharness,lin2026agentic}. Selection also
depends on the available supervision: RHO uses retrospective
self-validation and self-consistency followed by pairwise self-preference
\citep{pan2026evolving}, while HarnessEvolve diagnoses failures against
verified, answer-conditioned reference trajectories and gates updates
by quality and task performance \citep{jiang2026harnessevolve}.
% 中文对照：
% 在可执行 harness 层面，AFlow 将工作流构建表述为代码搜索问题 \citep{zhang2025aflow}。
% 一个关键区别是优化器可以访问哪些证据：
% Meta-Harness 开放历史代码、分数与原始轨迹，AHE 则将经验组织为分层证据集合，
% 并通过变更清单将编辑与预期作用关联起来 \citep{lee2026metaharness,lin2026agentic}。
% 候选选择也取决于可用的监督信号：
% RHO 使用回顾式自验证和自一致性分析，再通过成对自偏好进行选择 \citep{pan2026evolving}；
% HarnessEvolve 则对照经过验证、以答案为条件的参考轨迹诊断失败，
% 并根据质量与任务表现筛选更新 \citep{jiang2026harnessevolve}。

In embodied settings, Zetta
revises critics, recovery strategies, and tools \citep{ding2026zetta},
whereas SHAPER evolves skills and context-construction code with
context-aware diagnosis and prompts requesting local edits and preservation
of successful behavior \citep{wang2026shaper}. EnactOpt diagnoses adjacent
decisions by relating action feedback to the actual model input and
action at the next decision, while limiting editor-visible execution
feedback to structured summaries without raw-trajectory retrieval.
Unlike SHAPER's reported
skill-then-code schedule, EnactOpt permits either or both components to
change in one proposal, with prompts specifying six edit operators and
requesting supporting evidence and behavior to preserve.
% 中文对照：
% 在具身设置中，Zetta 修订检查器、恢复策略与工具 \citep{ding2026zetta}，
% SHAPER 则进化技能与上下文构造代码，结合考虑上下文的诊断，
% 并在提示中要求局部编辑和保留成功行为 \citep{wang2026shaper}。
% EnactOpt 将动作反馈与下一次决策时的实际模型输入和动作关联起来，对相邻决策进行诊断，
% 同时将编辑器可见的执行反馈限定为结构化总结，不提供原始轨迹检索。
% 不同于 SHAPER 所报告的先技能后代码的流程，EnactOpt 允许在同一次提案中修改一个或两个组件，
% 通过提示规定六类编辑算子，并要求提供支持证据与需保留的行为。

\section{Problem Setup}
\label{sec:scaffold-evolution}
% 中文标题：问题设定。

\subsection{Embodied Task Setting}
\label{sec:embodied-task-setting}
We consider an embodied task $q$ specified by an environment $\mathcal{E}_q$
with initial conditions, an instruction $\ell_q$, and a task-level success
criterion. At decision step $t$, the agent selects an action $a_t$ using
information drawn from $\ell_q$, the current observation $o_t$, and the
available interaction history $h_t$ of prior observations, actions, and
feedback. Executing $a_t$ yields the next observation $o_{t+1}$ and any
available execution feedback $f_{t+1}$; interaction continues until
termination. The agent is not assumed to observe the full environment
state or information reserved for task evaluation.
% 中文对照：
% 我们考虑一个具身任务 $q$，由带有初始条件的环境 $\mathcal{E}_q$、指令 $\ell_q$
% 和任务级成功判据共同规定。
% 在决策步骤 $t$，智能体利用从 $\ell_q$、当前观测 $o_t$ 和可用交互历史 $h_t$ 中选取的信息，
% 选择动作 $a_t$；$h_t$ 记录此前的观测、动作和反馈。
% 执行 $a_t$ 产生下一步观测 $o_{t+1}$ 以及可用的执行反馈 $f_{t+1}$，交互持续至终止。
% 不假设智能体能够观察到完整环境状态或专供任务评估使用的信息。

Let $\mathcal{M}$ denote the fixed model configuration used for task
execution and $\mathcal{R}$ the non-editable runtime responsible for
model-input assembly, model invocation, output parsing, and action
execution. Depending on the benchmark, $\mathcal{M}$ may include a
reasoning model coupled to a low-level policy, or a language model that
directly issues commands through a high-level action interface.
The index $t$ counts agent decisions rather than individual control steps,
so one action $a_t$ may invoke a low-level policy for multiple control
steps. Both $\mathcal{M}$ and $\mathcal{R}$ remain fixed during harness
evolution; benchmark-specific configurations are described in
Section~\ref{sec:experimental-setup}.
% 中文对照：
% 令 $\mathcal{M}$ 表示用于任务执行的固定模型配置，$\mathcal{R}$ 表示负责模型输入组装、
% 模型调用、输出解析和动作执行的不可编辑运行框架。
% 根据不同基准，$\mathcal{M}$ 可包含相互配合的推理模型和低层策略，
% 也可由通过高层动作接口直接发出命令的语言模型构成。
% $t$ 统计智能体决策而非单个控制步，因此一个动作 $a_t$ 可以调用低层策略执行多个控制步。
% $\mathcal{M}$ 和 $\mathcal{R}$ 在 harness 演化期间保持固定；
% 各基准的具体配置见第 \ref{sec:experimental-setup} 节。

\subsection{Execution Harness}
\label{sec:embodied-agent-scaffold}
We represent the editable part of the execution harness as
\begin{equation}
H = (S,C),
\label{eq:scaffold}
\end{equation}
where $S$ contains reusable textual skills that guide task execution,
exploration, and failure recovery. The executable context-construction
code $C$ selects, retains, links, and presents information derived from
available observations and interaction history. It processes observed
evidence rather than providing access to unobserved environment state.
Harness adaptation is restricted to $S$ and $C$; the model configuration
$\mathcal{M}$ and runtime $\mathcal{R}$ remain fixed.
% 中文对照：
% 我们将执行支架的可编辑部分表示为 $H=(S,C)$。
% 其中，$S$ 包含可复用的文本技能，用于指导任务执行、探索和失败恢复。
% 可执行的上下文构造代码 $C$ 对可用观测与交互历史中的信息进行选择、保留、关联与呈现。
% 它处理已经观察到的证据，而不是提供访问未观察环境状态的新渠道。
% harness 适应仅修改 $S$ 和 $C$；模型配置 $\mathcal{M}$ 与运行框架 $\mathcal{R}$ 保持固定。

At decision step $t$, we write the context produced by $C$ schematically as
\[
c_t = C(\ell_q,S,o_t,h_t).
\]
The fixed runtime $\mathcal{R}$ assembles $c_t$ with any additional
runtime-provided observations or messages to form $x_t$, the complete
input delivered to the decision-making model. Consequently, information
omitted from $c_t$ may still be present in $x_t$. Whether feedback was
presented to the model must therefore be assessed against $x_t$, rather
than the constructed context alone.
% 中文对照：
% 在决策步骤 $t$，我们将 $C$ 生成的上下文示意性地写为 $c_t=C(\ell_q,S,o_t,h_t)$。
% 固定运行框架 $\mathcal{R}$ 将 $c_t$ 与可能额外提供的观测或消息组装起来，
% 形成决策模型实际收到的完整输入 $x_t$。
% 因此，没有出现在 $c_t$ 中的信息仍可能出现在 $x_t$ 中。
% 判断反馈是否被呈现给模型，应检查 $x_t$，而不能只检查上下文构造代码的输出。

The artifacts $S$ and $C$ remain fixed throughout each rollout, whereas
$c_t$ and $x_t$ are constructed at each decision. For a task $q$, the fixed
execution system $(\mathcal{M},\mathcal{R})$ and harness $H$ induce a rollout
\begin{equation}
\xi \sim \operatorname{Rollout}
(\mathcal{E}_q,\ell_q;\mathcal{M},\mathcal{R},H),
\label{eq:scaffold-rollout}
\end{equation}
where $\xi$ denotes the recorded episode, including decision-level
observations, actual model inputs, actions, and returned feedback,
together with the task outcome reported at termination. How these
records are converted into feedback for the optimizer is specified in
Section~\ref{sec:local-patching}.
% 中文对照：
% 同一条 rollout 中，技能与代码产物 $S$、$C$ 保持不变，而 $c_t$ 和 $x_t$ 在每次决策时构造。
% 对于任务 $q$，固定执行系统 $(\mathcal{M},\mathcal{R})$ 与 harness $H$ 共同产生一条 rollout：
% $\xi \sim \operatorname{Rollout}(\mathcal{E}_q,\ell_q;\mathcal{M},\mathcal{R},H)$。
% $\xi$ 表示记录下来的 episode，包括决策级观测、实际模型输入、动作、返回的反馈，
% 以及终止时报告的任务结果。
% 如何将这些记录转换成优化器使用的反馈，见第 \ref{sec:local-patching} 节。

\subsection{Harness Adaptation Objective}
\label{sec:harness-adaptation-objective}
% 中文标题：执行支架适应目标。

With the model configuration $\mathcal{M}$ and runtime $\mathcal{R}$
fixed, adaptation optimizes only the harness $H=(S,C)$.
Let $\mathcal{H}$ denote the space of admissible harnesses that respect
the component boundaries and runtime-interface requirements defined above.
Let $p_{\mathrm{target}}$ be the target task distribution and $R_q(\xi)$
the task-level score assigned to a rollout $\xi$ of task $q$, with larger
values indicating better performance.
% 中文对照：
% 在模型配置 $\mathcal{M}$ 和运行框架 $\mathcal{R}$ 固定的情况下，
% 适应过程仅优化执行支架 $H=(S,C)$。
% 令 $\mathcal{H}$ 表示满足前述组件边界与运行接口要求的可行 harness 空间。
% 令 $p_{\mathrm{target}}$ 表示目标任务分布，$R_q(\xi)$ 表示任务 $q$ 的一次 rollout $\xi$
% 所获得的任务级得分，数值越大表示表现越好。

Our objective is to maximize the expected task score under
$p_{\mathrm{target}}$:
\begin{equation}
\begin{aligned}
J(H) &=
\mathbb{E}_{q \sim p_{\mathrm{target}}}
\mathbb{E}_{\xi \sim \operatorname{Rollout}(\mathcal{E}_q,\ell_q;\mathcal{M},\mathcal{R},H)}
\bigl[R_q(\xi)\bigr],\\
H^\star &\in \operatorname*{arg\,max}_{H \in \mathcal{H}} J(H).
\end{aligned}
\label{eq:harness-objective}
\end{equation}
For binary task-success scores, $J(H)$ is the expected success rate over
the target task distribution. Here, $H^\star$ denotes an ideal optimal
harness rather than the output guaranteed by the adaptation procedure.
% 中文对照：
% 我们的目标是最大化目标任务分布 $p_{\mathrm{target}}$ 下的期望任务得分：
% $J(H)=\mathbb{E}_{q\sim p_{\mathrm{target}}}
% \mathbb{E}_{\xi\sim\operatorname{Rollout}(\mathcal{E}_q,\ell_q;\mathcal{M},\mathcal{R},H)}[R_q(\xi)]$，
% $H^\star\in\operatorname*{arg\,max}_{H\in\mathcal{H}}J(H)$。
% 当任务得分是二值成功指标时，$J(H)$ 就是目标任务分布上的期望成功率。
% $H^\star$ 表示理想最优支架，而不是适应算法被保证能够找到的输出。

In practice, $J(H)$ is not directly observed during adaptation.
EnactOpt proposes revisions using finite execution feedback
collected on training tasks $\mathcal{D}_{\mathrm{train}}$ and selects
candidates according to empirical performance on validation tasks
$\mathcal{D}_{\mathrm{val}}$.
The target distribution may include task conditions absent from both
adaptation sets. Validation therefore provides a criterion for candidate
selection, while generalization beyond the adaptation conditions is
assessed separately.
% 中文对照：
% 在实际适应过程中，并不能直接观察到 $J(H)$。
% EnactOpt 利用训练任务 $\mathcal{D}_{\mathrm{train}}$ 上收集的有限执行反馈提出修订，
% 并依据候选在验证任务 $\mathcal{D}_{\mathrm{val}}$ 上的经验表现进行选择。
% 目标分布可能包含这两个适应集合都未覆盖的任务条件。
% 因此，验证表现用于候选选择，而超出适应条件的泛化能力需要另外评估。

\section{EnactOpt: Diagnosis-Guided Harness Evolution}
\label{sec:robopatch}
% 中文标题：EnactOpt：诊断引导的支架演化。

\begin{figure}[t]
\centering
\includegraphics[width=\linewidth]{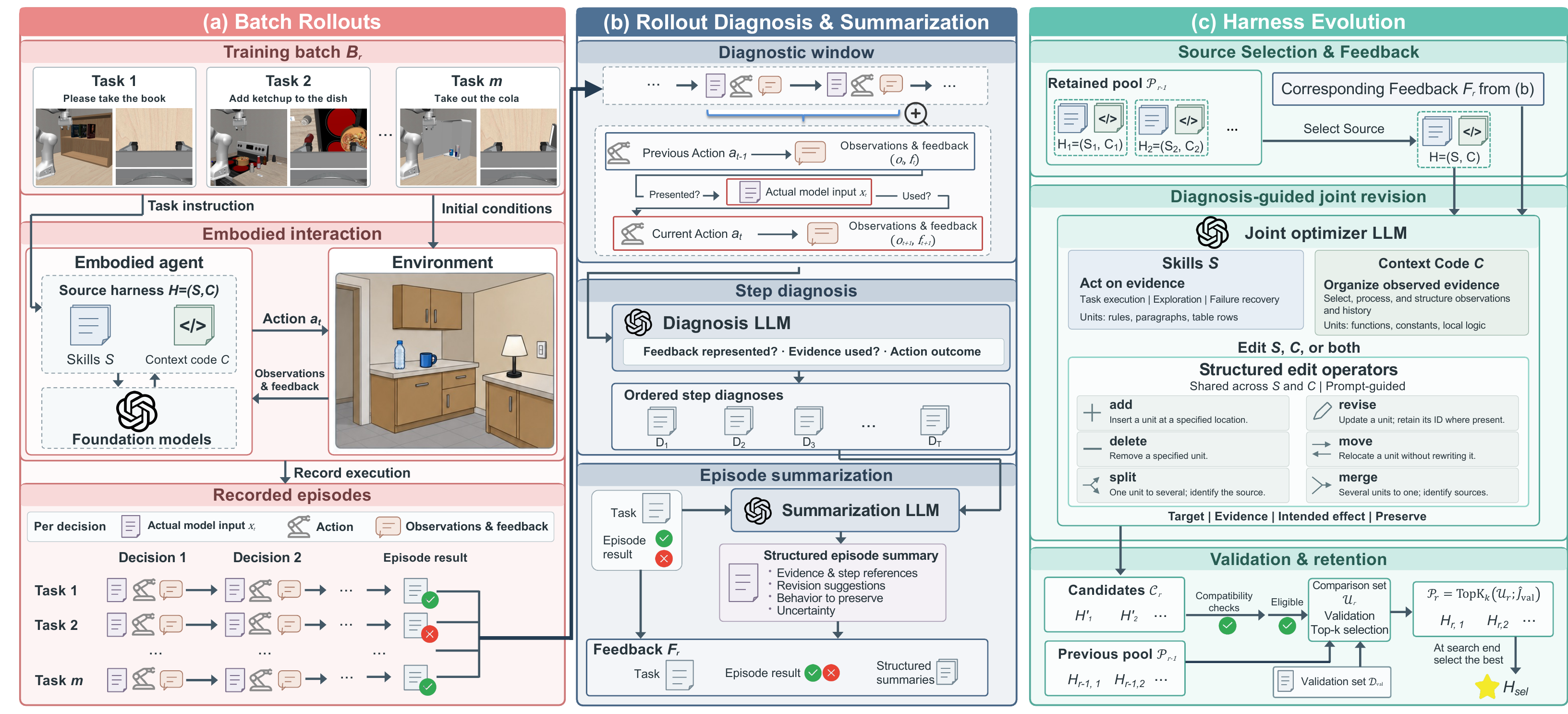}
\caption{Overview of EnactOpt.
(a) A source harness $H=(S,C)$ executes training batch $B_r$, recording
actual model inputs, actions, observations, and episode outcomes.
(b) Adjacent-decision diagnoses and episode summaries form its feedback $F_r$.
(c) A joint optimizer uses the same $H$ and $F_r$ to propose candidates
$\mathcal{C}_r$ through prompt-guided structured edit operators.
Compatible candidates are compared with the previous pool
$\mathcal{P}_{r-1}$ on validation tasks to form $\mathcal{P}_r$.
Scenes are illustrative, not experimental results.}
% 中文对照：
% EnactOpt 流程概览。
% (a) 来源支架 $H=(S,C)$ 执行训练批次 $B_r$，记录实际模型输入、动作、观测与 episode 结果。
% (b) 相邻决策诊断与 episode 总结构成该来源版本的反馈 $F_r$。
% (c) 联合优化器利用同一 $H$ 及其反馈 $F_r$，通过提示引导的结构化编辑算子提出候选 $\mathcal{C}_r$。
% 通过兼容性检查的候选与上一轮候选池 $\mathcal{P}_{r-1}$ 在验证任务上比较，形成 $\mathcal{P}_r$。
% 场景仅为示意，并非实验结果。
\label{fig:pipeline}
\end{figure}

\subsection{Evolution Overview}
\label{sec:evolution-overview}
\label{sec:single-pass-adaptation}
Figure~\ref{fig:pipeline} summarizes EnactOpt's iterative evolution loop.
Starting from $H_0=(S_0,C_0)$, EnactOpt maintains a pool of retained
harnesses while keeping $\mathcal{M}$ and $\mathcal{R}$ fixed.
At evolution iteration $r$, a source harness $H=(S,C)$ is selected from
$\mathcal{P}_{r-1}$ to execute a batch $B_r$ from the training set
$\mathcal{D}_{\mathrm{train}}$, producing rollouts and their native
execution outcomes.
% 中文对照：
% 图 \ref{fig:pipeline} 展示了 EnactOpt 的迭代演化过程。
% EnactOpt 从初始 harness $H_0=(S_0,C_0)$ 出发，在保持模型配置 $\mathcal{M}$
% 和运行框架 $\mathcal{R}$ 固定的情况下，维护一个保留版本池。
% 在第 $r$ 轮演化中，从 $\mathcal{P}_{r-1}$ 中选择来源 harness $H=(S,C)$，
% 执行训练集 $\mathcal{D}_{\mathrm{train}}$ 中的一个批次 $B_r$，产生 rollout 及其原生执行结果。

Each rollout is analyzed through adjacent-decision diagnoses, which are
aggregated into a structured episode summary. The task descriptions,
native outcomes, and complete summaries form the feedback $F_r$
associated with the source harness. A single joint optimizer receives
this same $H$ together with $F_r$ to propose candidate harnesses
$\mathcal{C}_r$, each comprising complete skill and context-code files.
Either component may remain unchanged in a proposal.
% 中文对照：
% 每条 rollout 通过相邻决策诊断进行分析，再将诊断汇总为结构化 episode 总结。
% 任务描述、原生执行结果和完整总结共同构成与来源 harness 对应的反馈 $F_r$。
% 单一联合优化器接收执行该批次的同一 $H$ 及其反馈 $F_r$，提出候选 harness 集合 $\mathcal{C}_r$，
% 其中每个候选都包含完整的技能与上下文代码文件。在一次提案中，任一组件都可以保持不变。

Candidates undergo local interface and runtime compatibility checks
before evaluation on $\mathcal{D}_{\mathrm{val}}$. Validation-based
selection updates the retained pool $\mathcal{P}_r$, which supplies source
versions for subsequent iterations. At the end of the search, a harness
$H_{\mathrm{sel}}$ is selected by validation performance and fixed for
test evaluation.
% 中文对照：
% 候选在验证集 $\mathcal{D}_{\mathrm{val}}$ 上接受评估之前，先经过本地接口与运行兼容性检查。
% 基于验证表现的选择更新保留池 $\mathcal{P}_r$，其中的版本为后续迭代提供来源。
% 搜索结束后，根据验证表现选出 harness $H_{\mathrm{sel}}$，并将其固定用于测试评估。

\subsection{Execution Diagnosis and Feedback Construction}
\label{sec:feedback-construction}
\label{sec:local-patching}
\label{sec:failure-diagnosis}
\paragraph{Adjacent-decision diagnosis.}
Task-level outcomes alone do not reveal whether relevant feedback was
missing from the model input or was available but inadequately used.
We therefore examine adjacent decisions (Figure~\ref{fig:pipeline}(b)).
At decision $t$, the diagnostic window relates the preceding action
$a_{t-1}$ and its returned observation and feedback $(o_t,f_t)$ to the
current model input $x_t$ and action $a_t$. It also includes the
observation and feedback $(o_{t+1},f_{t+1})$ returned by $a_t$ to assess
the current action's outcome. The first step has no preceding action
feedback; the final step remains assessable because no subsequent
decision is required.
% 中文对照：
% 相邻决策诊断。
% 仅凭任务级结果，无法判断相关反馈是没有进入模型输入，还是已经提供却未被合理利用。
% 因此，我们检查相邻决策（图 \ref{fig:pipeline}(b)）。
% 在决策 $t$，诊断窗口将前一步动作 $a_{t-1}$ 及其返回的观测与反馈 $(o_t,f_t)$，
% 与当前模型输入 $x_t$ 和动作 $a_t$ 关联起来。
% 窗口还包含 $a_t$ 返回的观测与反馈 $(o_{t+1},f_{t+1})$，用于评估当前动作的结果。
% 首步没有前序动作反馈；由于不需要后续决策，末步仍可接受评估。

The diagnostic LLM produces a report $D_t$ assessing whether preceding
feedback is faithfully presented in $x_t$, whether the current action
reflects appropriate use of that feedback, and what outcome the action
produces. The presentation check uses the full input $x_t$, not $c_t$
alone, because the runtime may append observations or messages.
Missing or misrepresented evidence may suggest a revision to $C$;
an inadequate response to available evidence may suggest a revision to
$S$. These are nonexclusive revision hypotheses, not causal proofs.
Ambiguous or incomplete evidence is retained as uncertainty.
% 中文对照：
% 诊断 LLM 生成报告 $D_t$，评估前序反馈是否被忠实地呈现在 $x_t$ 中、
% 当前动作是否体现了对该反馈的合理利用，以及该动作产生了什么结果。
% 呈现检查使用完整输入 $x_t$，而非仅检查 $c_t$，因为运行框架可能追加观测或消息。
% 证据缺失或错误呈现可能提示需要修订 $C$；对已有证据响应不当可能提示需要修订 $S$。
% 这些是非互斥的修订假设，而非因果证明。含糊或不完整的证据保留为不确定性。

\paragraph{Structured episode feedback.}
The episode summarizer receives the task description, native episode
outcome, and all ordered diagnoses $D_1,\ldots,D_T$, where $T$ is the
number of decisions; it does not receive the raw trajectory.
Its prompt requests a structured summary retaining concrete conditions,
observations and explanations reported by the diagnoses, and step
references for supporting and conflicting evidence. It also requests
repair suggestions, behavior to preserve, and unresolved uncertainty,
so that proposed changes account for useful existing behavior as well
as observed problems. The summarized evidence remains diagnosis-mediated:
step references do not independently verify an interpretation or
guarantee lossless summarization.
% 中文对照：
% 结构化 episode 反馈。
% episode 总结器接收任务描述、原生 episode 结果和全部有序诊断 $D_1,\ldots,D_T$，
% 其中 $T$ 为决策数；总结器不接收原始轨迹。
% 提示要求结构化总结保留具体条件、诊断报告中的观察与解释，以及支持和冲突证据的步骤引用。
% 提示还要求给出修复建议、需要保留的行为以及尚未消除的不确定性，
% 使拟议修改既考虑已观察到的问题，也顾及已有的有效行为。
% 总结中的证据仍经过诊断转述：步骤引用不能独立验证解释，也不能保证总结无损。

For a batch $B_r$ executed by the source harness $H$, the task
descriptions, native outcomes, and complete episode summaries form
$F_r$. The optimizer receives $H$ and $F_r$ without attached
raw-trajectory windows, actor replies, or context-history excerpts.
Step references indicate the evidence reported in the diagnoses, but
do not trigger raw-log retrieval. Raw logs remain available locally for
inspection, outside the optimizer's feedback interface. This feedback
guides the joint revision of $S$ and $C$ in
Section~\ref{sec:joint-revision}.
% 中文对照：
% 对于来源 harness $H$ 执行的批次 $B_r$，任务描述、原生结果和完整 episode 总结共同构成 $F_r$。
% 优化器接收 $H$ 与 $F_r$，不附带原始轨迹窗口、Actor 回复或上下文历史片段。
% 步骤引用标明诊断中报告的证据，但不会触发原始日志检索。
% 原始日志仍保留在本地供检查，不属于优化器的反馈接口。
% 这些反馈用于指导第 \ref{sec:joint-revision} 节中对 $S$ 与 $C$ 的联合修订。

\subsection{Joint Revision with Structured Edit Operators}
\label{sec:joint-revision}
\label{sec:edit-budgets}
Given $H=(S,C)$ and its batch feedback $F_r$, one joint optimizer LLM proposes
complete replacement files for both components:
\begin{equation}
H'=(S',C')=\operatorname{Optimizer}(S,C,F_r).
\label{eq:budgeted-edit}
\end{equation}
Each proposal yields a candidate $H'\in\mathcal{C}_r$, with either
component allowed to remain unchanged. Context-construction code
selects, processes, and organizes existing observations and interaction
history into task-relevant context, including structured state
representations. Skills guide task execution, further exploration, and
failure recovery. Both affect action selection; new environment
observations arise through agent actions or tool calls, not direct
environment interaction by the context builder.
% 中文对照：
% 给定 $H=(S,C)$ 及其批次反馈 $F_r$，单一联合优化器 LLM 为两个组件提出完整替换文件：
% $H'=(S',C')=\operatorname{Optimizer}(S,C,F_r)$。
% 每次提案产生一个候选 $H'\in\mathcal{C}_r$，其中任一组件都可以保持不变。
% 上下文构造代码选择、加工和组织已有观测与交互历史，形成与任务相关的上下文，
% 其中可以包括结构化状态表示。
% 技能指导任务执行、进一步探索与失败恢复。
% 两者都影响动作选择；新的环境观测来自智能体动作或工具调用，
% 而非上下文构造器直接与环境交互。

The optimizer is prompted to use six structured edit operators:
add, revise, delete, move, split, and merge. Editable units are Markdown
items, paragraphs, or table rows in skills, and functions, constants, or
local logic in context-construction code. For each proposed edit, the
optimizer is prompted to identify its target in the source harness, the
selected operator, supporting feedback, the intended effect, and behavior
to preserve. These prompt-level conventions guide incremental revision
while the optimizer returns complete files; preservation of untouched
content is requested rather than programmatically enforced.
% 中文对照：
% 提示要求优化器使用六类结构化编辑算子：新增、修订、删除、移动、拆分与合并。
% 技能的可编辑单元为 Markdown 条目、段落或表格行，上下文构造代码的可编辑单元为
% 函数、常量或局部逻辑。
% 对于每项拟议编辑，提示要求优化器指出其在来源 harness 中的修改对象、
% 所选算子、支持该修改的反馈、预期作用和需要保留的行为。
% 这些提示词层面的约定引导增量修订，而优化器仍输出完整文件；
% 未修改内容的保留由提示要求，而非由程序强制执行。

\subsection{Validation and Harness Retention}
\label{sec:validation-retention}
\label{sec:selection-memory}
Each candidate in $\mathcal{C}_r$ first undergoes local interface and
runtime compatibility checks under the fixed execution system.
Only compatible candidates proceed to evaluation on
$\mathcal{D}_{\mathrm{val}}$. Their empirical validation score
$\hat{J}_{\mathrm{val}}(H)$ is computed from the task-level scores
$R_q(\xi)$; for binary outcomes, it is the validation success rate.
Compatibility checks determine eligibility for evaluation, not
behavioral preservation or improved task performance.
% 中文对照：
% $\mathcal{C}_r$ 中的每个候选首先在固定执行系统下接受本地接口与运行兼容性检查。
% 只有通过检查的候选才进入验证集 $\mathcal{D}_{\mathrm{val}}$ 上的评估。
% 经验验证得分 $\hat{J}_{\mathrm{val}}(H)$ 根据任务级得分 $R_q(\xi)$ 计算；
% 当结果为二值时，该得分就是验证成功率。
% 兼容性检查决定候选是否具备评估资格，而不能证明行为得到保留或任务表现有所改善。

Let $\mathcal{U}_r$ denote the comparison set comprising compatible new
candidates and the previously retained harnesses $\mathcal{P}_{r-1}$.
Following the top-$k$ retention scheme in Figure~\ref{fig:pipeline}(c),
these versions are ranked by $\hat{J}_{\mathrm{val}}$, and up to $k$ harnesses are
retained:
\begin{equation}
\mathcal{P}_r = \operatorname{TopK}_k\!
\left(\mathcal{U}_r;\,\hat{J}_{\mathrm{val}}\right),
\label{eq:candidate-retention}
\end{equation}
where $k$ is the pool capacity and $\operatorname{TopK}_k$ selects the
highest-scoring versions. Thus, a new proposal competes with existing
versions rather than automatically replacing its source.
% 中文对照：
% 令 $\mathcal{U}_r$ 表示由通过兼容性检查的新候选与上一轮保留支架 $\mathcal{P}_{r-1}$ 组成的比较集合。
% 按照图 \ref{fig:pipeline}(c) 中的 top-$k$ 保留方式，这些版本依据
% $\hat{J}_{\mathrm{val}}$ 排序，并保留至多 $k$ 个支架：
% $\mathcal{P}_r=\operatorname{TopK}_k\!\left(\mathcal{U}_r;\,\hat{J}_{\mathrm{val}}\right)$，
% 其中 $k$ 为候选池容量，$\operatorname{TopK}_k$ 选出得分最高的版本。
% 因此，新提案需要与已有版本竞争，而不是自动替换其来源版本。

The updated pool $\mathcal{P}_r$ supplies source harnesses for subsequent
iterations under the experiment's search schedule. At the end of the
search, the retained harness with the highest validation score is
selected as $H_{\mathrm{sel}}$ and fixed for test evaluation.
Training feedback thus guides proposal generation, whereas validation
performance guides retention and final selection; performance beyond
the adaptation conditions is assessed separately at test time.
% 中文对照：
% 更新后的候选池 $\mathcal{P}_r$ 按照实验的搜索安排，为后续迭代提供来源支架。
% 搜索结束后，保留池中验证得分最高的支架被选为 $H_{\mathrm{sel}}$，并固定用于测试评估。
% 因此，训练反馈指导提案生成，验证表现指导候选保留与最终选择；
% 超出适应条件的表现则在测试阶段另行评估。

\section{Experiments}
\label{sec:experiments}

\begin{table}[!t]
\centering
\caption{Task success rates (\%, higher is better) across three benchmarks.
Bold indicates the best result in each column. Colored annotations show
percentage-point differences from Naive (green: positive; red: negative).
$\dagger$ denotes our skill-only implementations of TextGrad and Trace2Skill.}
% 中文对照：
% 三个基准上的任务成功率（\%，越高越好）。
% 粗体表示每列最高值，彩色标注表示相对 Naive 的百分点差（绿色为正，红色为负）。
% $\dagger$ 表示我们实现的仅修改技能的 TextGrad 与 Trace2Skill。
\label{tab:in-distribution-results}
\begin{tabularx}{\linewidth}{l*{3}{>{\centering\arraybackslash}X}}
\toprule[1pt]
\textbf{Method} & \textbf{VLABench} & \textbf{ALFWorld} & \textbf{EAI} \\
\midrule[0.7pt]
\rowcolor{mainheaderrow}
\multicolumn{4}{l}{\textbf{Reference Systems}} \\
Naive & \mainresult{40.87}{} & \mainresult{51.46}{} & \mainresult{68.09}{} \\
$H_0$ & \mainresult{47.39}{+6.52} & \mainresult{66.06}{+14.60} & \mainresult{72.34}{+4.26} \\
\midrule[0.4pt]
\rowcolor{mainheaderrow}
\multicolumn{4}{l}{\textbf{Adaptation Methods}} \\
TextGrad$^\dagger$ & \mainresult{36.09}{\textcolor{tableloss}{-4.78}} & \mainresult{86.50}{+35.04} & \mainresult{71.63}{+3.55} \\
Trace2Skill$^\dagger$ & \mainresult{40.54}{\textcolor{tableloss}{-0.33}} & \mainresult{78.83}{+27.37} & \mainresult{72.70}{+4.61} \\
GEPA & \mainresult{44.46}{+3.59} & \mainresult{83.21}{+31.75} & \mainresult{86.88}{+18.79} \\
\rowcolor{mainenactoptrow}
\textbf{EnactOpt} & \mainresult{\textbf{57.83}}{\textcolor{mainenactoptgain}{\textbf{+16.96}}} & \mainresult{\textbf{92.70}}{\textcolor{mainenactoptgain}{\textbf{+41.24}}} & \mainresult{\textbf{95.74}}{\textcolor{mainenactoptgain}{\textbf{+27.66}}} \\
\bottomrule[1pt]
\end{tabularx}
% 中文对照：
% 列：方法、VLABench Test920、ALFWorld Seen+Unseen、EAI Test282。
% 行：Naive、$H_0$、TextGrad、Trace2Skill、GEPA、EnactOpt。
% 分组：参照系统（Naive、$H_0$）；适配方法（TextGrad、Trace2Skill、GEPA、EnactOpt）。
% TextGrad 与 Trace2Skill 的 $\dagger$ 标记表示我们实现的仅修改技能的版本。
% 分组标题以冰川蓝底跨整表显示，文字左对齐并加粗。
% EnactOpt 行使用参考色块的浅粉红底，名称与主分数使用黑色粗体，增益使用深绿色粗体。
% 主分数居中，彩色下标标注相对 Naive 的百分点差。
% 正差值用绿色，负差值用红色。
% TextGrad 的 VLABench 最终 Test 成功数为 332/920（36.09\%），相对 Naive 为 -4.78 个百分点。
% TextGrad 的 ALFWorld 成功数为 237/274（86.50\%），相对 Naive 为 +35.04 个百分点。
% TextGrad 的 EAI 成功数为 202/282（71.63\%），相对 Naive 为 +3.55 个百分点。
% Trace2Skill 的 VLABench 成功数为 373/920（40.54\%），相对 Naive 为 -0.33 个百分点。
% Trace2Skill 的 ALFWorld 成功数为 216/274（78.83\%），相对 Naive 为 +27.37 个百分点。
% Trace2Skill 的 EAI 成功数为 205/282（72.70\%），相对 Naive 为 +4.61 个百分点。
% GEPA 的 ALFWorld 将缺失的一个 Unseen 实例计为失败，成功数为 228/274（83.21\%），
% 相对 Naive 为 +31.75 个百分点；EAI 成功率为 86.88\%，相对 Naive 为 +18.79 个百分点。
% $H_0$：+6.52、+14.60、+4.26；GEPA：+3.59、+31.75、+18.79；
% EnactOpt：+16.96、+41.24、+27.66。
\end{table}

\subsection{Experimental Setup}
\label{sec:experimental-setup}
\noindent\textbf{Benchmarks and protocol.}
We evaluate separately evolved harnesses on VLABench
\citep{zhang2025vlabench}, ALFWorld \citep{shridhar2020alfworld}, and
EAI's VirtualHome Action Sequencing \citep{li2024embodied}.
All MLLM components use GPT-5.4 mini \citep{openai2026gpt54};
VLABench additionally uses the benchmark-released $\pi_0$ policy
\citep{black2024pi_0}, fixed during evolution. ALFWorld and EAI use
high-level action interfaces. Train/validation/test sizes are 30/30/920
for VLABench and 30/30/282 for EAI.
For VLABench, adaptation and candidate selection use the in-distribution
track (T1); testing covers T1, commonsense-based target selection (T3),
context-rich instruction understanding (T4), and unseen textures (T6).
GEPA and EnactOpt share a frozen ALFWorld
30/30/274 split, with 140 Seen and 134 Unseen test instances.
GEPA's unscored Unseen instance is counted as a failure.
% 中文对照：
% 基准与协议。
% 我们在 VLABench \citep{zhang2025vlabench}、ALFWorld \citep{shridhar2020alfworld}
% 和 EAI 的 VirtualHome Action Sequencing \citep{li2024embodied} 上评估分别进化的 harness。
% 所有 MLLM 组件使用 GPT-5.4 mini \citep{openai2026gpt54}；
% VLABench 还使用基准发布的 $\pi_0$ 策略 \citep{black2024pi_0}，其参数在演化期间保持固定。
% ALFWorld 与 EAI 使用高层动作接口。VLABench 与 EAI 的训练/验证/测试规模
% 分别为 30/30/920 和 30/30/282。
% VLABench 的适配与候选选择使用分布内 track（T1）；测试覆盖 T1、
% 基于常识的目标选择（T3）、富含上下文的指令理解（T4）以及未见纹理（T6）。
% GEPA 与 EnactOpt 共用冻结的 ALFWorld 30/30/274 划分，测试包含 140 个 Seen 和 134 个 Unseen 实例。
% GEPA 缺失的一个 Unseen 实例计为失败。
% GEPA 计数备查（成功/全部实例）：Seen 112/140，Unseen 116/134，合计 228/274。

\noindent\textbf{Evolution settings.}
Motivated by the cost of collecting embodied interaction data, we
restrict EnactOpt to 30 training task instances per benchmark.
These tasks form three fixed, disjoint batches of ten, with one batch
per evolution iteration, yielding one training epoch.
Here, an epoch counts task coverage, not the total rollouts collected
by different source harnesses.
Candidates passing compatibility checks are evaluated on the full
30-instance validation set. The default pool capacity is $k=2$:
after each iteration, the top two harnesses among the new candidates
and the previous pool are retained by validation success rate.
After the third iteration, the retained harness with the highest
validation success rate is selected and fixed for test evaluation.
% 中文对照：
% 演化设置。
% 考虑到具身交互数据的获取成本，我们将每个基准上 EnactOpt 的训练任务实例数限制为 30。
% 这些任务组成三个固定且不重叠的批次，每批十个；每轮演化使用一个批次，三轮完成一个训练 epoch。
% 这里的 epoch 表示对训练任务的覆盖，而非不同来源 harness 产生的 rollout 总次数。
% 通过兼容性检查的候选在完整的 30 实例验证集上评估。
% 默认候选池容量为 $k=2$：每轮按验证成功率，从新候选与上一轮保留池中选出前两个 harness。
% 第三轮结束后，选择保留池中验证成功率最高的 harness，固定用于测试评估。

\noindent\textbf{Baselines and metrics.}
Naive uses each benchmark's official pipeline without our additional
harness; $H_0$ is the initial harness without adaptation.
TextGrad \citep{yuksekgonul2024textgrad} and Trace2Skill
\citep{ni2026trace2skill} are our skill-only implementations, with
context-construction code kept fixed.
GEPA \citep{agrawal2026gepa} and EnactOpt start from $H_0$.
We report task success rates (\%) and differences in percentage points.
These comparisons are not matched-compute evaluations. Benchmark-specific
data protocols are described in \appendixref{app:evolution-details}.
% 中文对照：
% 对比方法与指标。
% Naive 使用各基准的官方流程，不加入我们额外的 harness；$H_0$ 是未经适配的初始 harness。
% TextGrad \citep{yuksekgonul2024textgrad} 和 Trace2Skill \citep{ni2026trace2skill}
% 是我们实现的仅修改技能的版本，上下文构造代码保持固定。
% GEPA \citep{agrawal2026gepa} 与 EnactOpt 从 $H_0$ 出发。
% 我们报告任务成功率（\%），并以百分点表示差值。
% 这些比较并非等计算量评估。
% 各基准的数据协议见附录 \ref{app:evolution-details}。
\ifdefined\enactoptarxiv
The evolution procedure is described in Section~\ref{sec:robopatch},
and \appendixref{app:harness-artifacts} provides the evolved skills
and context-construction code for each benchmark.
% 中文对照（仅 arXiv）：
% 第 \ref{sec:robopatch} 节描述演化流程，
% 附录 \ref{app:harness-artifacts} 提供各基准进化后的技能与上下文构造代码。
\fi

\noindent\textbf{Optimization cost.}
For the VLABench ablations, we count complete optimization passes:
candidate generation, evaluation on the full validation set, and pool
retention for one batch. EnactOpt and the diagnosis and edit-operator
ablations use one pass per batch; separate $S$/$C$ revision uses two,
one per component. The three-batch schedule therefore requires three
versus six passes, normalized to $1\times$ and $2\times$ relative to
EnactOpt; $H_0$ is not optimized.
% 中文对照：
% 优化成本。
% 对 VLABench 消融，我们统计完整优化循环的次数；
% 每次循环包含针对一个批次的候选生成、完整验证集评估和候选池保留。
% EnactOpt、诊断消融和编辑算子消融对每批执行一次循环；
% S/C 分开修订对每批执行两次，分别对应一个组件。
% 因此，三个批次共需三次或六次循环，以 EnactOpt 为基准归一化为 $1\times$ 或 $2\times$；
% $H_0$ 不进行优化。

\subsection{Main Results}
\label{sec:main-results}
With 30 training task instances per benchmark, EnactOpt improves on
the initial harness across both low-level policy execution and
high-level action interfaces (Table~\ref{tab:in-distribution-results}).
Its gains over $H_0$ are 10.43, 26.64, and 23.40 percentage points
on VLABench, ALFWorld, and EAI, respectively.
These gains extend beyond the benefit of constructing $H_0$:
execution experience supports further refinement of an already
configured agent without changing its underlying models.
% 中文对照：
% 每个基准使用 30 个训练任务实例时，EnactOpt 在低层策略执行与高层动作接口两类设置中
% 均优于初始 harness（表 \ref{tab:in-distribution-results}）。
% 相对 $H_0$，VLABench、ALFWorld 和 EAI 上的增益分别为 10.43、26.64 和 23.40 个百分点。
% 这些收益超出了构造初始 $H_0$ 本身的作用：
% 在不改变底层模型的情况下，执行经验仍能支持对已有执行配置的进一步改进。

EnactOpt also outperforms every evaluated adaptation baseline, but
the relative strengths of those baselines vary by benchmark.
Our skill-only TextGrad implementation is the strongest adaptation
baseline on ALFWorld, whereas GEPA leads the alternatives on VLABench
and EAI. On VLABench, however, even GEPA's adaptation from $H_0$
falls below the unchanged $H_0$ reference. Thus, the ability to revise
model-external artifacts does not by itself ensure a better harness.
These complete-pipeline comparisons motivate examining the diagnostic
and revision choices in Section~\ref{sec:ablations}, rather than
attributing the gains to editable components alone.
% 中文对照：
% EnactOpt 也优于全部已评估的适配基线，但这些基线的相对优势随基准变化。
% 我们实现的仅修改技能的 TextGrad 在 ALFWorld 的适配基线中最强，
% GEPA 则在 VLABench 和 EAI 的其他适配方法中领先。
% 然而，在 VLABench 上，即使从 $H_0$ 开始适配的 GEPA 也低于未修改的 $H_0$ 参照。
% 因此，允许修订模型外部产物，本身并不保证获得更好的 harness。
% 这些完整流程的比较促使我们在第 \ref{sec:ablations} 节进一步考察诊断与修订设计，
% 而非仅将收益归因于组件可编辑。

\subsection{Generalization across Test Conditions}
\label{sec:generalization}
\begin{figure}[t]
\centering
\includegraphics[width=\linewidth]{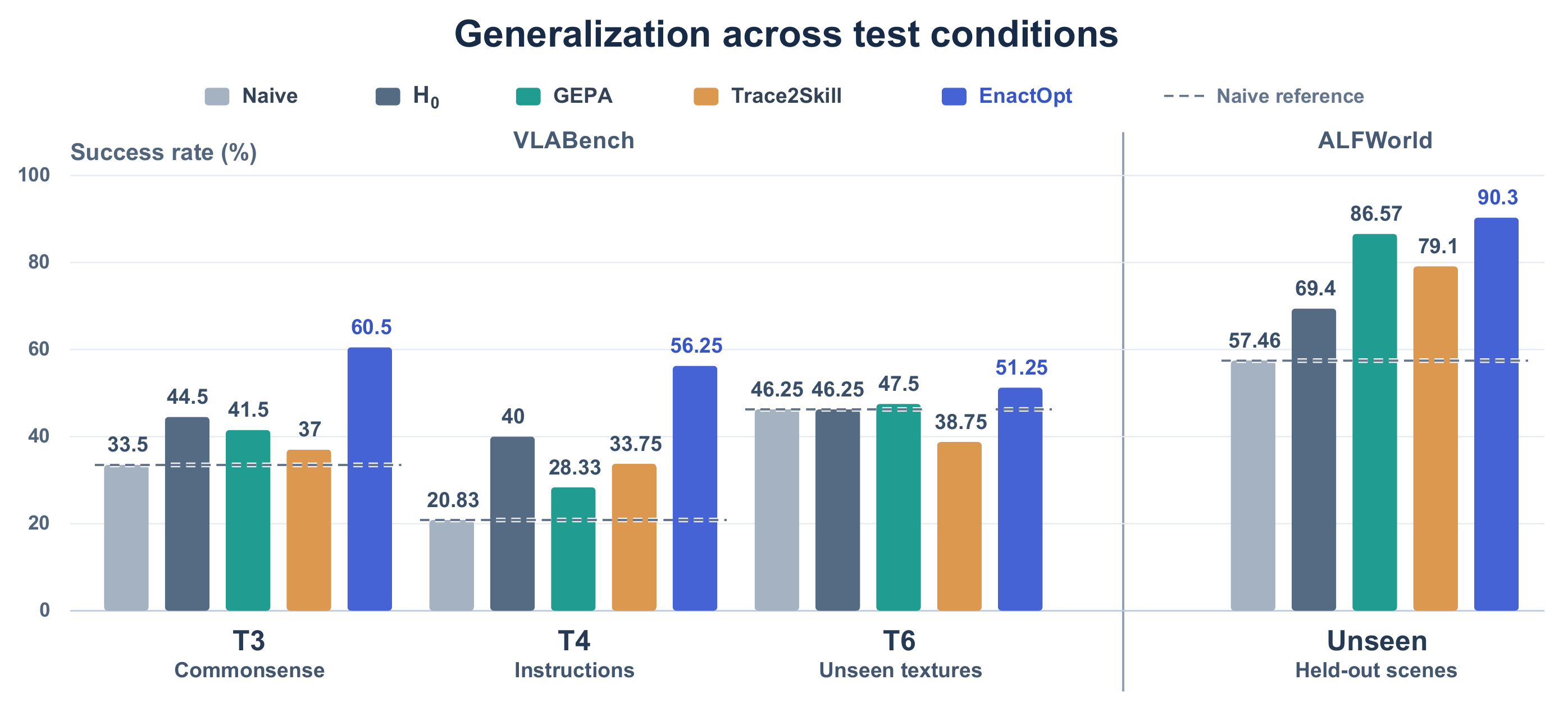}
\caption{Generalization on VLABench T3/T4/T6 and ALFWorld Unseen.
Bars show task success rates (\%); horizontal dashed lines denote
the Naive baseline. Adaptation is performed separately for each benchmark.
The plotted results come from the same evaluation runs as
Table~\ref{tab:in-distribution-results}, which reports overall success
rates on the full test sets.}
% 中文对照：
% VLABench T3/T4/T6 与 ALFWorld Unseen 上的泛化表现。
% 柱子表示任务成功率（\%），水平虚线表示 Naive 基线。各基准分别进行适配。
% 图中结果来自与表 \ref{tab:in-distribution-results} 相同的评测运行；
% 此处展示部分测试条件的分组结果，主表报告完整测试集上的总体成功率。
\label{fig:generalization}
\end{figure}

Adaptation on VLABench T1 transfers to other test conditions, but the
gains vary with the type of shift (Figure~\ref{fig:generalization}).
Although evolution and candidate selection use only T1, EnactOpt
improves T3 and T4 over $H_0$ by 16.00 and 16.25 percentage points,
compared with 5.00 points on T6; it also exceeds GEPA on all three tracks.
The revisions thus remain useful beyond the adaptation conditions,
with the largest improvements on the tracks that change how the
target or request must be interpreted, rather than visual appearance.
% 中文对照：
% 在 VLABench T1 上适配所得的收益能延伸至其他测试条件，但幅度随条件变化的类型而不同
% （图 \ref{fig:generalization}）。
% 尽管演化与候选选择只使用 T1，EnactOpt 在 T3 和 T4 上相对 $H_0$ 分别提高
% 16.00 和 16.25 个百分点，在 T6 上提高 5.00 个百分点；三个 track 上也均优于 GEPA。
% 因此，这些修订在适配条件之外仍然有用，
% 且最大改善出现在改变目标或请求理解方式的 track，而非改变视觉外观的 track。

ALFWorld complements this comparison with a change in scene rather
than evaluation track. On Unseen scenes, EnactOpt exceeds both $H_0$
and GEPA, with a 3.73-percentage-point advantage over GEPA in the runs
shown. The advantage therefore persists against an adapted baseline,
not only the initial harness. Unseen retains the same task families;
both benchmarks assess within-benchmark transfer without test-time
harness revision, rather than transfer of one harness across benchmarks.
% 中文对照：
% ALFWorld 通过场景变化而非评估 track 的变化，为上述比较提供补充。
% 在图中所用的 Unseen 场景运行中，EnactOpt 同时优于 $H_0$ 和 GEPA，
% 相对 GEPA 的优势为 3.73 个百分点。
% 因而优势不仅相对初始 harness 存在，相对经过适配的基线也仍然存在。
% Unseen 保留相同的任务类型；两项评估均是在测试时不再修订 harness 的条件下
% 考察基准内部迁移，而非将同一个 harness 迁移至其他基准。

\subsection{Ablation Studies}
\label{sec:ablations}
% Keep the ablation table off Figure 3's page.
\suppressfloats[t]
We examine diagnostic evidence, structured edit operators, and joint
revision on VLABench (Table~\ref{tab:vlabench-ablation}).
\emph{w/o Adjacent-Decision Diagnosis} retains joint revision and edit
operators but uses only task descriptions and terminal outcomes as
diagnostic evidence. Relative to $H_0$, it gains 3.75 percentage points
on In-Distribution but loses 1.91 points on Generalization, whereas
EnactOpt improves both groups. This contrast shows why in-distribution
improvement alone is not a sufficient measure of a reusable revision:
the gain from outcome-only feedback does not extend to the shifted
test conditions.
% 中文对照：
% 我们在 VLABench 上考察诊断证据、结构化编辑算子与联合修订（表 \ref{tab:vlabench-ablation}）。
% 去掉相邻决策诊断的变体保留联合修订和编辑算子，但仅使用任务描述与最终结果作为诊断证据。
% 相对 $H_0$，该变体在 In-Distribution 上提高 3.75 个百分点，
% 但在 Generalization 上降低 1.91 个百分点，而 EnactOpt 在两组上均有提升。
% 这一对照说明，仅凭分布内改善不足以衡量修订的可复用性：
% 仅结果反馈所带来的收益并未延伸至条件变化后的测试集。

\emph{w/o Structured Edit Operators} retains diagnosis and joint revision
but removes prompt-level editing conventions. Its success rates fall
below $H_0$ in both groups, by 2.08 percentage points In-Distribution
and 2.50 points on Generalization. Retaining diagnostic feedback is
therefore not sufficient to obtain useful revisions in this comparison.
The result points to the importance of how the optimizer uses that
evidence to organize edits, not only whether it receives the evidence.
% 中文对照：
% 去掉结构化编辑算子的变体保留诊断和联合修订，但移除提示词层面的编辑约定。
% 其成功率在两组中均低于 $H_0$：In-Distribution 低 2.08 个百分点，
% Generalization 低 2.50 个百分点。
% 因此，在本次比较中，仅保留诊断反馈并不足以形成有用的修订。
% 结果提示，重要的不仅是优化器是否接收证据，还包括它如何利用证据组织修改。

\emph{w/o Joint Revision} still edits both $S$ and $C$, but separately.
Its In-Distribution success remains close to EnactOpt
(63.33\% vs.\ 63.75\%), whereas Generalization is lower
(50.88\% vs.\ 55.74\%), despite using twice as many optimization passes.
The comparison concerns how revisions are proposed and evaluated,
not whether both components can change. Additional component-wise
passes do not recover the generalization performance of joint revision
in these runs.
% 中文对照：
% 去掉联合修订的变体仍修改 S 和 C，但分别优化两者。
% 其 In-Distribution 成功率仍与 EnactOpt 接近（63.33\% 对 63.75\%），
% 但即使用了两倍的优化循环，其 Generalization 仍较低（50.88\% 对 55.74\%）。
% 这一比较关注的是如何提出和评估修订，而不是是否允许两个组件都发生变化。
% 在这些运行中，额外的逐组件优化循环未能达到联合修订的泛化表现。

\begin{table}[t]
\centering
\caption{VLABench ablations on T1 (In-Distribution) and T3/T4/T6
(Generalization). Rates (\%); subscripts: percentage-point changes
from $H_0$. Cost: optimization passes relative to EnactOpt.}
% 中文对照：
% VLABench 在 T1（In-Distribution）与 T3/T4/T6（Generalization）上的消融结果。
% 分数为成功率（\%），下标表示相对 $H_0$ 的百分点差；
% 成本统计完整优化循环的次数，并以 EnactOpt 为基准归一化。
\label{tab:vlabench-ablation}
\begin{tabularx}{\linewidth}{l*{2}{>{\centering\arraybackslash}X}c}
\toprule[1pt]
\textbf{Variant} & \textbf{In-Distribution} & \textbf{Generalization} & \textbf{Optimization cost} \\
\midrule[0.7pt]
$H_0$ & \mainresult{58.33}{} & \mainresult{43.53}{} & -- \\
w/o Adjacent-Decision Diagnosis & \mainresult{62.08}{+3.75} & \mainresult{41.62}{\textcolor{tableloss}{-1.91}} & $1\times$ \\
w/o Structured Edit Operators & \mainresult{56.25}{\textcolor{tableloss}{-2.08}} & \mainresult{41.03}{\textcolor{tableloss}{-2.50}} & $1\times$ \\
w/o Joint Revision & \mainresult{63.33}{+5.00} & \mainresult{50.88}{+7.35} & $2\times$ \\
\rowcolor{mainenactoptrow}
\textbf{EnactOpt} & \mainresult{\textbf{63.75}}{\textcolor{mainenactoptgain}{\textbf{+5.42}}} & \mainresult{\textbf{55.74}}{\textcolor{mainenactoptgain}{\textbf{+12.21}}} & $1\times$ \\
\bottomrule[1pt]
\end{tabularx}
% 中文对照：
% 列：配置、In-Distribution、Generalization、相对优化成本。
% 行：$H_0$、去掉相邻决策诊断、去掉结构化编辑算子、去掉联合修订、EnactOpt。
% EnactOpt 行使用与表 1 相同的粉红底，增益使用深绿色小字。
% 增益由成功计数计算后再四舍五入；诊断与编辑算子消融的相对优化成本均为 1 倍，
% 去掉联合修订为 2 倍，EnactOpt 为 1 倍；H0 不进行 harness 优化。
% 去掉结构化编辑算子：T1 135/240，T3 83/200，T4 81/240，T6 115/240；
% Generalization 为 279/680（41.03\%），Overall 为 414/920（45.00\%）。
\end{table}

\section{Discussion}
\label{sec:discussion}
The results point to a distinction between the capabilities of an
underlying model and the way an agent uses those capabilities.
Skills and context-construction code can change how observed evidence
is organized and acted upon while model parameters and runtime interfaces
remain fixed. This makes the execution harness a complementary
adaptation target, rather than a substitute for model or policy training.
Improvements at this level do not demonstrate new perception or
low-level control capabilities.
% 中文对照：
% 这些结果提示，应区分底层模型具有的能力与智能体使用这些能力的方式。
% 在模型参数与运行接口保持固定的情况下，技能和上下文构造代码可以改变
% 已观察证据的组织方式，以及智能体依据这些证据采取行动的方式。
% 因此，执行 harness 是一种互补的适配对象，而非模型或策略训练的替代品。
% 这一层面的改善并不证明获得了新的感知或低层控制能力。

The harness artifacts in \appendixref{app:harness-artifacts} show concrete
changes in memory use, recovery, and controller-facing instructions.
ALFWorld expands the retained-history window from two to 50 turns and
revises search and responses to ineffective actions. EAI replaces
stopping after one correction with continued attempts within budget,
while emphasizing current state and the latest feedback.
VLABench explicitly presents observed entity names and prescribes short,
canonical commands that preserve their exact spelling.
These are observed differences between harnesses, not isolated
attributions of the performance gains.
% 中文对照：
% 附录 \ref{app:harness-artifacts} 中的 harness 文本与代码展示了记忆使用、恢复策略
% 和面向控制器的指令形式等方面的具体变化。
% ALFWorld 将保留历史的窗口从两轮扩展至 50 轮，并调整搜索和无效动作后的响应。
% EAI 从一次纠错后停止改为在预算内继续尝试，同时强调当前状态与最新反馈。
% VLABench 显式呈现已经观察到的实体名称，并要求使用保留精确拼写的简短规范命令。
% 这些是 harness 之间可观察到的差异，而不是对性能增益的孤立因素归因。

Our evaluation focuses on within-benchmark adaptation.
\ifdefined\enactoptarxiv
These simulated and text-based evaluations do not establish safe
operation on physical robots; deploying evolved execution code would
require additional safety validation and oversight.
% 中文对照（仅 arXiv）：
% 这些仿真与文本环境中的评估不能证明真实机器人上的安全运行；
% 部署进化后的执行代码需要额外的安全验证与监督。
\fi
Future work
could investigate the transfer of evolved harnesses across benchmarks,
their application to broader task compositions, and their reliability
on physical robots. Controlled, matched-budget studies on independent
test data would help disentangle the contributions of skill revision,
context-code revision, and diagnostic feedback. Measuring both
evolution and execution costs would also clarify the trade-off between
task performance and overall resource use.
% 中文对照：
% 本文的评估聚焦于基准内部的适配。
% 后续工作可以研究演化所得 harness 在不同基准之间的迁移、
% 对更广泛任务组合的适用性，以及在真实机器人上的可靠性。
% 在独立测试数据上开展受控且计算预算匹配的实验，将有助于区分技能修订、
% 上下文代码修订与诊断反馈各自的贡献。
% 同时衡量演化与执行阶段的成本，也有助于明确任务表现与总体资源开销之间的权衡。

\section{Conclusion}
\label{sec:conclusion}
We presented EnactOpt, which uses execution diagnosis to jointly evolve
skills and context-construction code while keeping the underlying models
fixed. Evaluations on VLABench, ALFWorld, and EAI show improved task
success, with gains also observed under the evaluated within-benchmark
shifts. The VLABench ablations show that variants with similar
in-distribution success can generalize differently; joint revision also
achieves higher generalization success with fewer optimization passes
than separate updates. These findings motivate treating evidence
presentation and action guidance as a coupled adaptation target, and
assessing harness revisions by whether they remain useful beyond the
conditions used for evolution.
% 中文对照：
% 我们提出了 EnactOpt，利用执行诊断联合演化技能与上下文构造代码，同时保持底层模型固定。
% 在 VLABench、ALFWorld 和 EAI 上的评估显示任务成功率有所提升，
% 并在已评估的基准内部条件变化下取得收益。
% VLABench 消融表明，分布内成功率相近的配置，其泛化表现仍可能不同；
% 联合修订也以更少的优化循环取得了比分开更新更高的泛化成功率。
% 这些发现支持将证据呈现与行动指导作为相互耦合的适配对象，
% 并以修订能否在演化所用条件之外继续发挥作用来衡量其价值。

\ifdefined\enactoptarxiv
\subsection*{Acknowledgments}

Generative AI tools, including GitHub Copilot, assisted with manuscript
drafting and editing, literature lookup, citation formatting, LaTeX
tables, and figure-preparation code. The authors remain responsible
for the final text, claims, and research artifacts.
% 中文对照：
% 包括 GitHub Copilot 在内的生成式 AI 工具辅助了稿件起草与编辑、文献查找、
% 引用格式整理、LaTeX 表格及图示制作代码。
% 作者仍对最终文本、论断和研究产物负责。
\else
\subsection*{AI use statement}

Generative AI tools, including GitHub Copilot, assisted with manuscript
drafting and language polishing, literature lookup and citation formatting,
LaTeX tables, and code for figure preparation. Model use in the evaluated
agents is described in Section~\ref{sec:experimental-setup}. The authors
retain responsibility for the final text, claims, and research artifacts.
% 中文对照：
% 包括 GitHub Copilot 在内的生成式 AI 工具辅助了稿件起草与语言润色、文献查找与引用格式整理、
% LaTeX 表格，以及用于图示制作的代码。
% 被评估智能体中的模型使用情况见第 \ref{sec:experimental-setup} 节。
% 作者对最终文本、论断和研究产物负责。

\subsection*{Ethics statement}

Our experiments use official VLABench, ALFWorld, and EAI benchmark
resources in simulated or text-based environments. The results do not
establish safe operation on physical robots; deploying evolved execution
code would require additional safety validation and oversight.
% 中文对照：
% 我们的实验使用 VLABench、ALFWorld 和 EAI 的官方基准资源，在仿真或文本环境中进行。
% 这些结果不能证明真实机器人上的安全运行；部署进化后的执行代码需要额外的安全验证与监督。

\subsection*{Reproducibility statement}

Section~\ref{sec:robopatch} describes the harness evolution procedure.
Section~\ref{sec:experimental-setup} and
\appendixref{app:evolution-details} document the evaluated task subsets,
data partitions, and training-batch construction.
\appendixref{app:harness-artifacts} provides the evolved skill texts
and context-construction code for each benchmark.
% 中文对照：
% 第 \ref{sec:robopatch} 节描述 harness 演化流程。
% 第 \ref{sec:experimental-setup} 节与附录 \ref{app:evolution-details}
% 记录所评估的任务子集、数据划分和训练批次构造。
% 附录 \ref{app:harness-artifacts} 提供各基准进化后的技能文本与上下文构造代码。
\fi

\bibliography{references}
\bibliographystyle{iclr2027_conference}

\clearpage
\appendix
\etocdepthtag.toc{appendix}
\section*{Appendix}
\label{app:contents}

% 附录目录仅列出 A、B 等节及其编号小节，正式内容从下一页开始。
\begingroup
\hypersetup{hidelinks,linktoc=all}
\etocsettocstyle{\subsection*{Contents}}{}
\etocsettagdepth{main}{none}
\etocsettagdepth{appendix}{subsection}
\tableofcontents
\endgroup
\clearpage

\section{EnactOpt Implementation and Evaluation Details}
\label{app:details}
\label{app:evolution-details}
% 中文标题：EnactOpt 实现与评估细节。
This appendix records implementation and evaluation details.
% 中文对照：
% 本附录记录实现与评估细节。

\subsection{VLABench Protocol}
We use six primitive tasks:
\texttt{select\_chemistry\_tube}, \texttt{select\_poker},
\texttt{select\_drink}, \texttt{add\_condiment},
\texttt{select\_book}, and \texttt{select\_toy}.
The VLABench-released \texttt{pi0-primitive-10task}
checkpoint\footnote{\url{https://huggingface.co/VLABench/pi0-primitive-10task}}
remains frozen during harness evolution.
Task selection was informed by the checkpoint's reported T1 performance
and T1--T4 performance gaps, rather than random sampling of primitive tasks.
% 中文对照：
% VLABench 协议。
% 所用的六个 primitive 任务为：
% \texttt{select\_chemistry\_tube}（选试管）、\texttt{select\_poker}（选扑克牌）、
% \texttt{select\_drink}（选饮料）、\texttt{add\_condiment}（添加调料）、
% \texttt{select\_book}（选书）和 \texttt{select\_toy}（选玩具）。
% 采用 VLABench 发布的 \texttt{pi0-primitive-10task} 检查点，发布地址见脚注；
% 其参数在 harness 进化期间保持冻结。
% 任务类别的选择参考了该检查点已报告的 T1 表现和 T1--T4 表现差距，
% 而非从 primitive 任务中随机抽样。

\begin{figure}[p]
\centering
\includegraphics[page=1,width=\linewidth]{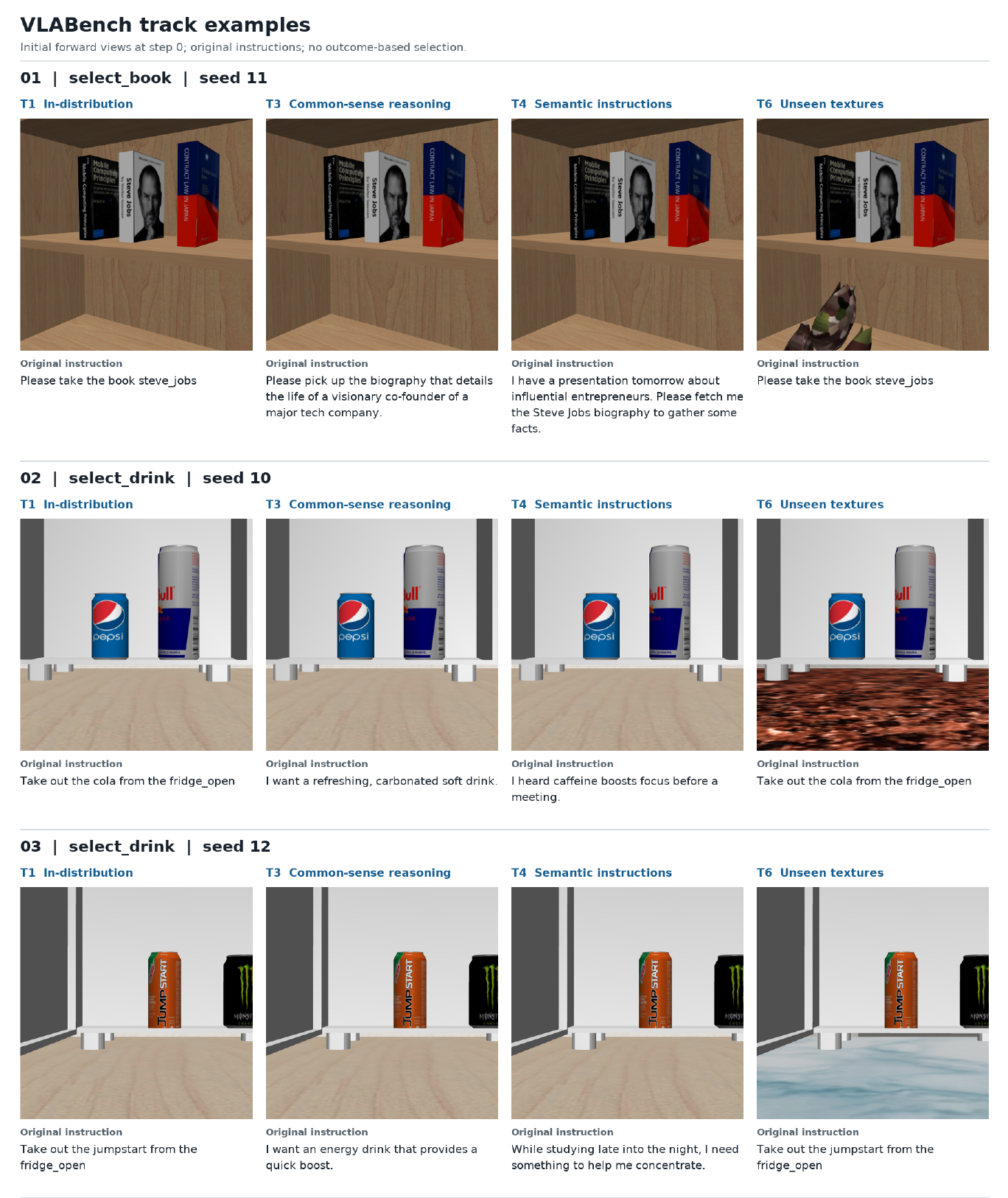}
\caption{VLABench track examples for book and drink selection.
Each row compares initial forward views and task instructions across
T1, T3, T4, and T6. T3/T4 illustrate changes in the instruction,
while T6 illustrates changes in the visual setting.
The images show initial states, not execution outcomes.}
% 中文对照：
% VLABench 书籍与饮料选择任务的 track 示例。
% 每行对照 T1、T3、T4、T6 的初始前视图及任务指令。
% T3/T4 展示指令的变化，T6 展示视觉环境的变化。
% 图片呈现初始状态，而非执行结果。
\label{fig:vlabench-track-examples}
\end{figure}

\begin{figure}[p]
\centering
\includegraphics[page=2,width=\linewidth]{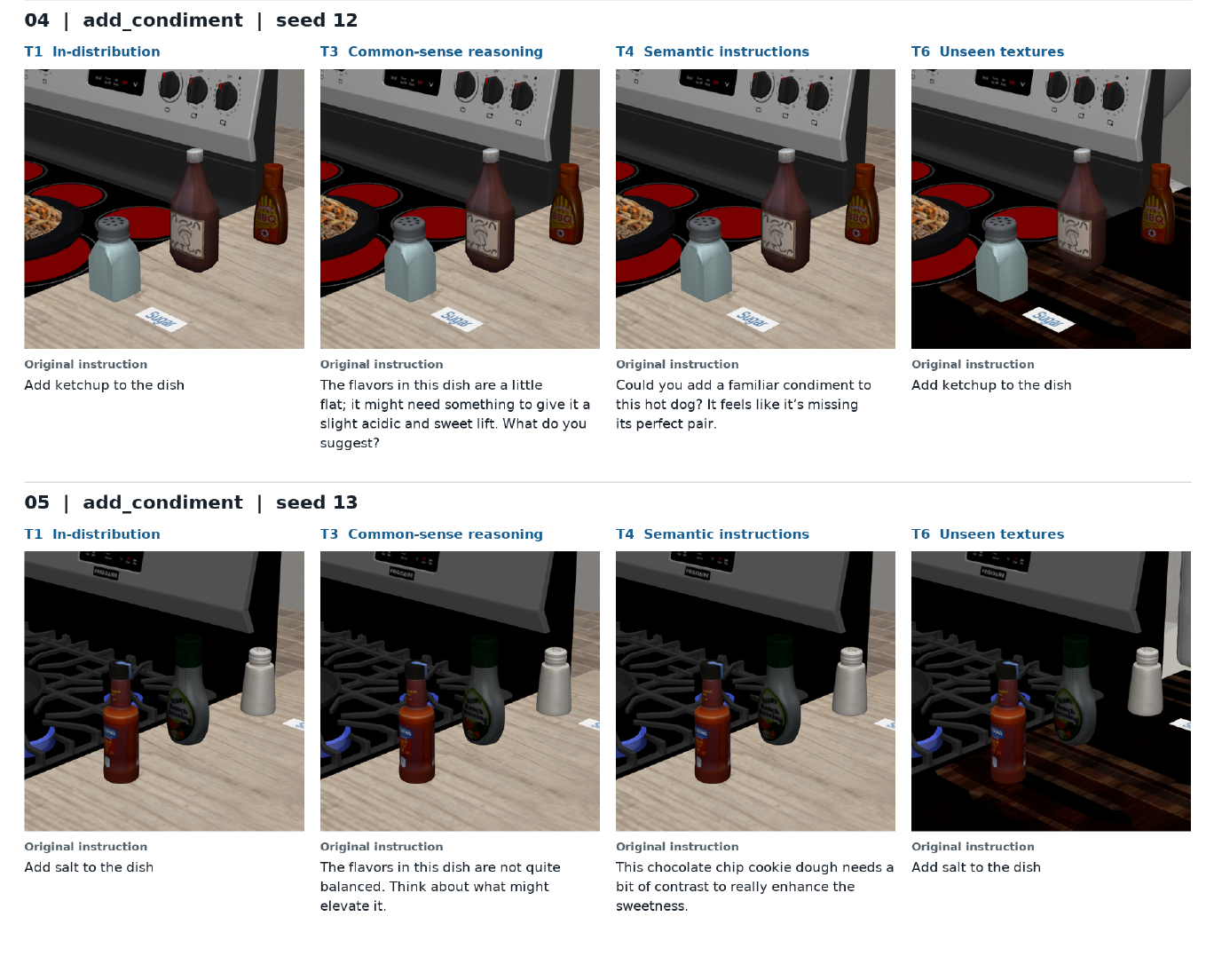}
\caption{VLABench track examples for condiment addition.
The columns follow the same T1/T3/T4/T6 order as
Figure~\ref{fig:vlabench-track-examples}, showing initial forward views
and task instructions for two further episodes.}
% 中文对照：
% VLABench 调料添加任务的 track 示例。
% 各列沿用图 \ref{fig:vlabench-track-examples} 的 T1/T3/T4/T6 顺序，
% 展示另外两个 episode 的初始前视图及任务指令。
\label{fig:vlabench-track-examples-condiments}
\end{figure}

We use the official T1 (In-Distribution), T3 (Common Sense),
T4 (Semantic Instruction), and T6 (Unseen Texture)
tracks.\footnote{\url{https://github.com/OpenMOSS/VLABench\#evaluate}}
T3 and T4 concern commonsense target identification and context-rich
instruction understanding, respectively; their reasoning demands can
overlap. T6 changes textures, materials, and backgrounds; it is not a
texture-only, single-factor intervention.
Figures~\ref{fig:vlabench-track-examples} and
\ref{fig:vlabench-track-examples-condiments} illustrate these tracks
with initial observations and task instructions.
% 中文对照：
% 使用官方 T1（分布内）、T3（常识）、T4（语义指令）和 T6（未见纹理）四个 track，说明见脚注。
% T3 和 T4 分别涉及常识目标识别与富含上下文信息的指令理解，两者的推理需求可能重叠。
% T6 同时改变纹理、材质和背景，并非只改变纹理的单因素干预。
% 图 \ref{fig:vlabench-track-examples} 和 \ref{fig:vlabench-track-examples-condiments}
% 通过初始观测与任务指令展示这四个 track 的例子。

We partition the official task--track episode lists using the fixed
indices in Table~\ref{tab:vlabench-splits}. Train30 and Val30 use T1 only.
Test920 contains 240, 200, 240, and 240 episodes from T1, T3, T4, and T6,
respectively. T3 excludes \texttt{select\_poker}; T2 is not included.
% 中文对照：
% 按表 \ref{tab:vlabench-splits} 的固定索引划分官方任务与 track 对应的 episode 列表。
% Train30 和 Val30 只使用 T1。Test920 中 T1、T3、T4、T6 分别有 240、200、240、240 个 episode。
% T3 不包含 \texttt{select\_poker}，测试范围不包含 T2。

\begin{table}[htbp]
\centering
\caption{Fixed VLABench episode splits. Indices are zero-based and
inclusive within each official task--track list. Counts include only
valid selected task--track pairs.}
% 中文对照：
% 固定的 VLABench episode 划分。索引在每个官方任务与 track 对应的列表内从零开始，区间包含两端。
% 数量仅统计所选任务中有效的任务与 track 组合。
\label{tab:vlabench-splits}
\begin{tabular*}{\linewidth}{@{\extracolsep{\fill}}llrr@{}}
\hline
Use & Track & Indices & Episodes \\
\hline
Harness evolution (Train30) & T1 & 0--4 & 30 \\
Candidate selection (Val30) & T1 & 5--9 & 30 \\
Test & T1, T3, T4, T6 & 10--49 & 920 \\
\hline
\end{tabular*}
% 中文对照：
% 列：用途、track、索引、episode 数量。
% 行：Harness 进化（Train30），T1，每类索引 0--4，共 30 局；
% 候选选择（Val30），T1，每类索引 5--9，共 30 局；
% 测试，T1、T3、T4、T6，每个有效任务与 track 组合的索引 10--49，共 920 局。
\end{table}

For EnactOpt, Train30 is partitioned into three fixed, disjoint Train10
batches, each containing all six tasks with one or two episodes per task.
Their union is Train30, but this membership partition does not bound
the total number of training rollouts. Val30 is used for candidate selection.
% 中文对照：
% 对 EnactOpt，Train30 划分为三个固定且不重叠的 Train10 批次，
% 每批包含全部六类任务，每类一至两个 episode。
% 三个批次的并集为 Train30，但成员划分并不约束训练 rollout 的总次数。
% Val30 用于候选选择。

Episode identities remain fixed across iterations, but simulator
realizations and policy actions need not be deterministic.
Exclusion from harness evolution does not establish exclusion from policy pretraining
or prior fine-tuning.
% 中文对照：
% Episode 身份在迭代间保持固定，但仿真实现与策略动作不一定是确定性的。
% 未参与 harness 进化，并不能证明这些 episode 未出现在策略预训练或此前微调数据中。

\clearpage

\subsection{ALFWorld Protocol}
\label{app:alfworld-protocol}
We use ALFWorld 0.4.2 with \texttt{AlfredTWEnv} and all six supported
task families. Following the text-environment loader, we exclude
unsupported movable and sliced variants and require valid game files
and \texttt{solvable=true}. The filtered official pools contain 3,553
training, 140 Seen, and 134 Unseen instances. Eligibility uses dataset
annotations, not screening by agent success.
% 中文对照：
% ALFWorld 协议。
% 我们使用 ALFWorld 0.4.2 的 \texttt{AlfredTWEnv}，保留全部六种受支持的任务类型。
% 按文本环境加载器的规则，排除不支持的 movable 和 sliced 变体，
% 并要求有效游戏文件及 \texttt{solvable=true}；可解标记缺失或为假的实例均不纳入。
% 过滤后的官方任务池包含 3,553 个训练实例、140 个 Seen 实例和 134 个 Unseen 实例。
% 可用性依据数据集标注确定，而非根据智能体是否成功进行筛选。

Train30 and Val30 are selected from the official training pool.
We hash-order its 108 scenes with seed 42 and divide them into two
disjoint pools of 54 scenes. Within each pool, game identifiers are
hash-ordered separately for each task family using the same seed,
and the first five instances per family are selected
(Table~\ref{tab:alfworld-splits}). This deterministic, task-stratified
procedure yields 30 instances per partition. Train30 and Val30 span
23 and 20 scenes, respectively, share neither instances nor scenes,
and remain fixed throughout harness evolution.
% 中文对照：
% Train30 和 Val30 均从官方训练池中选取。
% 我们按种子为 42 的哈希顺序排列其中的 108 个场景，划分为两个不相交的 54 场景池。
% 在每个场景池内，对各任务类型的游戏标识符分别使用相同种子进行哈希排序，
% 每类取前五个实例（表 \ref{tab:alfworld-splits}）。
% 这一确定性的任务分层选取过程为每个开发集产生 30 个实例。
% Train30 和 Val30 分别覆盖 23 个和 20 个场景，实例和场景均不重叠，
% 并在 harness 进化期间保持固定。

\begin{table}[htbp]
\centering
\caption{Task-family counts for the fixed ALFWorld Train30/Val30 split
and the complete Seen/Unseen evaluation sets.}
% 中文对照：
% 冻结的 ALFWorld Train30/Val30 划分与完整 Seen/Unseen 评估集的各任务类型实例数。
\label{tab:alfworld-splits}
\begin{tabular*}{\linewidth}{@{\extracolsep{\fill}}lrrrr@{}}
\hline
Task family & Train30 & Val30 & Seen & Unseen \\
\hline
Pick \& Place & 5 & 5 & 35 & 24 \\
Pick Two \& Place & 5 & 5 & 24 & 17 \\
Examine in Light & 5 & 5 & 13 & 18 \\
Clean \& Place & 5 & 5 & 27 & 31 \\
Heat \& Place & 5 & 5 & 16 & 23 \\
Cool \& Place & 5 & 5 & 25 & 21 \\
\hline
Total & 30 & 30 & 140 & 134 \\
\hline
\end{tabular*}
% 中文对照：
% 列：任务类型、Train30、Val30、最终 Seen、最终 Unseen。
% 行：拾取并放置、拾取两个物体并放置、在光照下检查、清洁后放置、加热后放置、冷却后放置。
% 每类在 Train30 和 Val30 中各有五个实例；各列合计为 30、30、140 和 134。
\end{table}

We partition the fixed Train30 into three disjoint Train10 batches by
task-stratified shuffling with \texttt{random.Random(42)}. Each batch
contains one or two instances per family, and the three batches cover
Train30 exactly once. Training rollouts supply harness-revision feedback
without updating model parameters; Val30 is used for candidate selection.
% 中文对照：
% 我们使用 \texttt{random.Random(42)} 按任务类型分层打乱，
% 将固定的 Train30 划分为三个不重叠的 Train10 批次。
% 每批每类包含一至两个实例，三个批次恰好覆盖 Train30 一遍。
% 训练执行轨迹提供 harness 修订反馈，不更新模型参数；Val30 用于候选选择。

The selected harness is fixed for evaluation on all 140 official
\texttt{valid\_seen} and 134 \texttt{valid\_unseen} instances, without
further subsampling. These sets are instance-disjoint from Train30
and Val30, but Seen shares some scenes with the development sets.
Unseen denotes scenes absent from the official training pool.
% 中文对照：
% 所选 harness 固定后，在官方 \texttt{valid\_seen} 的全部 140 个实例及
% \texttt{valid\_unseen} 的全部 134 个实例上评估，不再抽样。
% 这两个集合与 Train30、Val30 的实例不重叠，但 Seen 与开发集存在部分场景重叠。
% Unseen 指官方训练池中未出现的场景。

\subsection{EAI Protocol}
We use all 342 VirtualHome Action Sequencing instances in
\texttt{scene\_1}, covering 26 official activity categories.
All partitions come from this same scene; no instances are excluded
based on model performance.
% 中文对照：
% EAI 协议。
% 我们使用 \texttt{scene\_1} 中 VirtualHome Action Sequencing 的全部 342 个实例，
% 覆盖 26 个官方活动类别。所有划分均来自同一场景，不按模型表现筛选或剔除实例。

We form indivisible groups by taking the transitive closure of two
relations: a shared numeric identifier prefix, or equivalence of
symbolic initial-state graphs and public goal specifications under
attributed directed-graph isomorphism. The comparison uses the full
symbolic graphs with geometric attributes removed; reference action
sequences are not used. This yields 310 groups, comprising 278
singletons and 32 pairs, while retaining all 342 instances.
Each group is assigned wholly to one partition.
% 中文对照：
% 我们对两种关联关系取传递闭包，形成不可拆分的实例组：
% 数字标识符前缀相同，或符号初始状态图与公开目标经带属性有向图同构判定为等价。
% 比较使用去除几何属性后的完整符号图，不使用参考动作序列。
% 最终得到 310 个组，包括 278 个单例组和 32 个双例组；全部 342 个实例仍被保留。
% 每个组整体分配到同一集合。

Each of the 23 categories with at least three instances receives a
quota of one training and one validation instance. The seven most
frequent eligible categories receive a second instance in each
development partition, with frequency ties broken by label.
Each partition therefore contains $16\times1+7\times2=30$ instances
covering 23 categories. The official categories ``Brush teeth''
(two instances), ``Get some water'' (one), and ``Take shower'' (two)
are assigned entirely to testing. These quotas prioritize category
coverage rather than proportional sampling.
% 中文对照：
% 对至少有三个实例的 23 个类别，训练集和验证集的基础配额各为一个实例。
% 实例最多的七类在每个开发集中再增加一个实例，频数相同时按标签排序。
% 因此每个开发集包含 $16\times1+7\times2=30$ 个实例，覆盖 23 类。
% 官方类别 ``Brush teeth''（两个实例）、``Get some water''（一个）和
% ``Take shower''（两个）全部分配至测试集。这些配额优先保证类别覆盖，而非等比例抽样。

Following a seed-42 hash ordering of groups, we try assigning each
group to Train, then Val, then Test. We commit the first choice for
which an integer-programming feasibility check confirms that the
remaining groups can still satisfy the remaining category quotas.
All instances outside Train30 and Val30 form Test282, with no
additional test subsampling. Table~\ref{tab:eai-splits} summarizes
the fixed partition.
% 中文对照：
% 按种子为 42 的哈希顺序处理各组，依次尝试将当前组分配至 Train、Val、Test。
% 通过整数规划检查剩余组能否继续满足剩余类别配额，并采用首个可行选择。
% 训练集与验证集之外的全部实例构成 Test282，不再额外抽取测试子集。
% 表 \ref{tab:eai-splits} 汇总了冻结的划分。

\begin{table}[htbp]
\centering
\caption{Frozen custom EAI split for VirtualHome Action Sequencing in
\texttt{scene\_1}. Category counts are distinct official activity labels
within each partition.}
% 中文对照：
% VirtualHome Action Sequencing 在 \texttt{scene\_1} 上冻结的自建 EAI 划分。
% 类别数指各集合内不同的官方活动标签数量。
\label{tab:eai-splits}
\begin{tabular*}{\linewidth}{@{\extracolsep{\fill}}lrrl@{}}
\hline
Partition & Instances & Categories & Use \\
\hline
Train & 30 & 23 & Harness evolution \\
Val & 30 & 23 & Candidate selection \\
Test & 282 & 26 & Evaluation \\
\hline
Total & 342 & 26 & All \texttt{scene\_1} instances \\
\hline
\end{tabular*}
% 中文对照：
% 列：划分、实例数、类别数、用途。
% 行：Train，30 个实例、23 类，用于 harness 进化；
% Val，30 个实例、23 类，用于候选选择；
% Test，282 个实例、26 类，用于评估；
% 总计，342 个实例、26 类，包含 \texttt{scene\_1} 的全部实例。
\end{table}

Train30 contains only singleton groups. We divide it deterministically
into three disjoint Train10 batches, each containing ten distinct
activity labels. Their union is Train30. These counts describe
instance membership, not the total number of training rollouts.
% 中文对照：
% Train30 中均为单例组。我们将其确定性地划分为三个不重叠的 Train10 批次，
% 每批包含十个不同的活动标签，三个批次的并集为 Train30。
% 这些计数描述实例的成员划分，而非训练执行轨迹的总次数。

\clearpage
\section{Benchmark Harness Artifacts}
\label{app:harness-artifacts}
% 中文标题：各基准的 Harness 文本与代码。

\subsection{VLABench}
\label{app:vlabench-harness-artifacts}

\subsubsection{Evolved Harness}
\label{app:vlabench-evolved}
The following boxes reproduce the planner skill and context-construction
code obtained after EnactOpt evolution for VLABench.
% 中文对照：
% 以下文本框保留 EnactOpt 在 VLABench 上进化得到的规划技能与上下文构造代码。

\begin{harnesslisting}{VLABench (EnactOpt): Planner Skill}
# Primitive Planner Skill

You are the high-level planner for a **primitive** VLABench manipulation task.
Each episode is a single short manipulation, not a long multi-stage plan.

## What you see
Each round you receive:
- the task instruction
- the current camera views
- the prior execution history
- the scene spellings in `entities`
- the current state, including what is held and the end-effector pose

## Your job
Emit exactly **one** short command for the low-level controller.

The controller is frozen and weak at paraphrase. It works best on short, canonical
commands that preserve the benchmark’s exact target and container tokens.

## Core rule
Use the shortest command that still preserves the task’s exact target and final
desired state.

Do **not**:
- invent new object names
- normalize token-like labels into space-separated words
- add extra spatial qualifiers
- add strategy text or explanations
- repeat a failed command unchanged if the visible scene has changed

## Preserve exact scene spellings
When the scene provides a token-like label, copy that spelling exactly.
Keep underscores, capitals, and container suffixes exactly as shown in `entities`
and in the task instruction.

Examples of exact spellings:
- `contract_law_in_japan`
- `CuCl2`
- `jumpstart`
- `fridge_open`
- `giftbox_seen`
- `ace_of_hearts`

Do not rewrite these into natural language forms.

## Canonical command forms
Prefer these literal patterns when they match the task:

- `Add <condiment> to the dish`
- `Please take the book <title>`
- `Take out the <chemical> solution`
- `Take out the <drink> from the fridge_open`
- `Please pick the poker <card>`
- `Put the <toy> into the giftbox_seen`

If the instruction already matches one of these forms, keep the command equally
short and literal, preserving the exact target token.

## Poker tasks
For poker episodes, the card name must be the exact scene entity spelling,
such as `ace_of_hearts` or `7_of_hearts`.
Do not write card names with spaces.

Example:
- task/instruction: `Please pick the poker ace of hearts`
- preferred subgoal: `Please pick the poker ace_of_hearts`

## History use
Look at the latest scene and history to decide whether the task is already done,
partially done, or still untouched.

- If the target is untouched, issue the direct canonical command.
- If the target is already grasped or removed, keep the same final-state goal but
  do not add new narration.
- If a previous attempt failed and the visible scene changed, adapt only by
  choosing the still-missing primitive outcome.

Do not blindly repeat a failed sentence if the scene has moved on.

## Style
- one command only
- no bullets in the subgoal
- no commentary inside the command
- no mention of cameras, state, or history

## Good behavior
- stay literal
- preserve exact benchmark tokens
- use the smallest command that can still be executed
- avoid paraphrases that change the object spelling
\end{harnesslisting}

\clearpage
% 长代码框略收紧行距，避免返回字典的末几行单独占页。
\begin{harnesslisting}[9.2]{VLABench (EnactOpt): Context Code}
"""Context builder for primitive VLABench manipulation tasks.

This version keeps the context compact and grounded:
- preserve literal task wording
- surface current state facts (held objects, ee pose, entities, track)
- summarize history briefly instead of replaying everything verbatim
- attach only the most recent completed-round frames plus live views
- keep live forward/right/wrist images last so truncation preserves them
"""

from __future__ import annotations

CAMERA_ORDER = ("forward", "right", "wrist")
MAX_HISTORY_TEXT_ROUNDS = 4
MAX_HISTORY_IMAGE_ROUNDS = 2


def _map_get(obj, key, default=None):
    if obj is None:
        return default
    try:
        getter = obj.get
    except Exception:
        getter = None
    if getter is not None:
        try:
            return getter(key, default)
        except Exception:
            pass
    try:
        return obj[key]
    except Exception:
        return default


def _short(text, limit=120):
    if text is None:
        return ""
    try:
        s = str(text)
    except Exception:
        s = repr(text)
    s = s.replace("\n", " ").strip()
    if len(s) <= limit:
        return s
    return s[: max(0, limit - 1)].rstrip() + "…"


def _fmt_num(value):
    try:
        return f"{float(value):.3f}"
    except Exception:
        return _short(value, 20)


def _iter_items(value):
    if value is None:
        return []
    if isinstance(value, str):
        return [value]
    if hasattr(value, "tolist") and not isinstance(value, (list, tuple, set, dict)):
        try:
            value = value.tolist()
        except Exception:
            pass
    if isinstance(value, dict):
        items = []
        for k, v in value.items():
            if isinstance(k, str):
                items.append(k)
            elif isinstance(v, str):
                items.append(v)
        if items:
            return items
        return [str(value)]
    try:
        return list(value)
    except Exception:
        return [value]


def _format_sequence(value, limit_items=8, item_limit=40):
    items = []
    seen = set()
    for item in _iter_items(value):
        s = _short(item, item_limit)
        if not s:
            continue
        if s in seen:
            continue
        seen.add(s)
        items.append(s)
        if len(items) >= limit_items:
            break
    if not items:
        return ""
    extra = 0
    try:
        total = len(list(_iter_items(value)))
        extra = max(0, total - len(items))
    except Exception:
        extra = 0
    text = ", ".join(items)
    if extra:
        text += f", …(+{extra})"
    return text


def _format_pose(ee_pose):
    if ee_pose is None:
        return ""
    try:
        if hasattr(ee_pose, "tolist"):
            ee_pose = ee_pose.tolist()
    except Exception:
        pass
    if isinstance(ee_pose, (list, tuple)) and ee_pose:
        vals = list(ee_pose)
        if len(vals) >= 4:
            pos = ", ".join(_fmt_num(v) for v in vals[:3])
            grip = _fmt_num(vals[-1])
            return f"pos=[{pos}], gripper={grip}"
        return "[" + ", ".join(_short(v, 20) for v in vals) + "]"
    return _short(ee_pose, 80)


def _first_nonempty(*values):
    for v in values:
        if v is None:
            continue
        if isinstance(v, str) and not v.strip():
            continue
        if v == {} or v == [] or v == ():
            continue
        return v
    return None


def _round_summary(index, record):
    planner = _map_get(record, "planner_output", {}) or {}
    parts = []

    sent = _first_nonempty(
        _map_get(record, "actor_subgoal", None),
        _map_get(planner, "subgoal", None),
    )
    if sent is not None:
        parts.append(f"sent={_short(sent, 120)!r}")

    steps_used = _first_nonempty(
        _map_get(record, "actor_steps_used", None),
        _map_get(record, "steps_used", None),
    )
    step_budget = _first_nonempty(
        _map_get(record, "actor_step_budget", None),
        _map_get(record, "step_budget", None),
    )
    if steps_used is not None:
        if step_budget is not None:
            parts.append(f"steps={steps_used}/{step_budget}")
        else:
            parts.append(f"steps={steps_used}")

    status = _first_nonempty(
        _map_get(record, "status", None),
        _map_get(record, "actor_status", None),
        _map_get(record, "execution_status", None),
    )
    runtime_error = _first_nonempty(
        _map_get(record, "runtime_error", None),
        _map_get(record, "error", None),
    )
    if status is not None:
        parts.append(f"status={_short(status, 40)}")
    if runtime_error is not None:
        parts.append(f"error={_short(runtime_error, 80)}")

    progress = _first_nonempty(_map_get(planner, "progress", None))
    if progress is not None:
        parts.append(f"progress={_short(progress, 120)!r}")

    if not parts:
        parts.append("no recorded details")
    return f"r{index}: " + "; ".join(parts)


def _get_camera_frame(images, camera):
    frame = _map_get(images, camera, None)
    if frame is not None:
        return frame
    try:
        return getattr(images, camera)
    except Exception:
        return None


def build_context(task: str, history: list[dict], observation, skill: str, llm=None) -> dict:
    del skill  # read-only planner prompt is already selected; keep builder deterministic.
    del llm    # no evidence-based benefit for extra summarization here.

    history = history or []
    extra = getattr(observation, "extra", None) or {}
    state = getattr(observation, "state", None) or {}
    live_images = getattr(observation, "images", None) or {}

    task_text = (task or _map_get(observation, "instruction", "") or "").strip()
    if not task_text:
        task_text = "No task text provided."

    track = _map_get(extra, "track", None)
    entities_raw = _map_get(extra, "entities", None)
    grasped_raw = _map_get(state, "grasped", None)
    ee_pose_raw = _map_get(state, "ee_pose", None)

    entities_text = _format_sequence(entities_raw, limit_items=12, item_limit=50)
    held_text = _format_sequence(grasped_raw, limit_items=8, item_limit=50)
    ee_pose_text = _format_pose(ee_pose_raw)

    # Build current-state text first.
    state_lines = []
    if track is not None:
        state_lines.append(f"track: {_short(track, 50)}")
    if held_text:
        state_lines.append(f"held: {held_text}")
    else:
        state_lines.append("held: none reported")
    if ee_pose_text:
        state_lines.append(f"ee_pose: {ee_pose_text}")
    if entities_text:
        state_lines.append(f"scene entities: {entities_text}")

    # Summarize history compactly.
    history_lines = []
    if not history:
        history_lines.append("No completed rounds yet.")
    else:
        show_from = max(0, len(history) - MAX_HISTORY_TEXT_ROUNDS)
        if show_from > 0:
            history_lines.append(f"{show_from} earlier completed rounds omitted from text to reduce stale context.")
        for idx in range(show_from, len(history)):
            rec = history[idx] or {}
            history_lines.append(_round_summary(idx, rec))

    # Attach images: most recent completed rounds first, then live views last.
    images = []
    tags = []

    frame_history = _map_get(extra, "frame_history", None) or []
    if not isinstance(frame_history, list):
        try:
            frame_history = list(frame_history)
        except Exception:
            frame_history = []

    paired_rounds = min(len(frame_history), len(history))
    start_round = max(0, paired_rounds - MAX_HISTORY_IMAGE_ROUNDS)
    for idx in range(start_round, paired_rounds):
        frame_dict = frame_history[idx] or {}
        for camera in CAMERA_ORDER:
            frame = _get_camera_frame(frame_dict, camera)
            if frame is None:
                continue
            images.append(frame)
            tags.append(f"round{idx}_after_{camera}")

    live_present = []
    for camera in CAMERA_ORDER:
        frame = _get_camera_frame(live_images, camera)
        if frame is None:
            continue
        images.append(frame)
        tags.append(f"now_{camera}")
        live_present.append(camera)

    text_sections = [
        "# Task\n" + task_text,
        "# Current state\n" + "\n".join(state_lines),
        "# History\n" + "\n".join(history_lines),
        "# Views\n"
        + (
            "Attached images are ordered from older history to the live scene; the live forward/right/wrist views are last."
            if images
            else "No camera images available."
        ),
    ]

    hints = []
    if track is not None:
        hints.append(f"track={_short(track, 60)}")
    if held_text:
        hints.append(f"held={held_text}")
    else:
        hints.append("held=none")
    if entities_text:
        hints.append(f"entities={entities_text}")
    if not history:
        hints.append("fresh round")
    hints.append("use exact scene spellings")
    hints.append("live views are last")

    metadata = {
        "builder": "context_compact_state_v1",
        "task": task_text,
        "track": track,
        "history_rounds": len(history),
        "history_text_rounds_shown": len(history_lines) - (1 if len(history) > MAX_HISTORY_TEXT_ROUNDS else 0),
        "history_image_rounds_shown": paired_rounds - start_round,
        "images_attached": len(images),
        "current_cameras": live_present,
        "held": _iter_items(grasped_raw) if grasped_raw is not None else [],
        "ee_pose": ee_pose_text,
        "entities_preview": _iter_items(entities_raw)[:12] if entities_raw is not None else [],
    }

    return {
        "text": "\n\n".join(text_sections),
        "images": images,
        "image_tags": tags,
        "hints": hints,
        "metadata": metadata,
    }
\end{harnesslisting}

\clearpage
\subsection{ALFWorld}
\label{app:alfworld-harness-artifacts}

\subsubsection{Evolved Harness}
\label{app:alfworld-evolved}
The following boxes reproduce the final evolved ALFWorld skill and
context code. The skill's original \texttt{H0} heading is retained verbatim.
% 中文对照：
% 以下文本框保留 ALFWorld 最终进化版本的技能与上下文代码。
% 技能原文标题中的 \texttt{H0} 按原样保留，两个框均对应进化后配置。

\begin{harnesslisting}[9.2]{ALFWorld (EnactOpt): Skill}
# ALFWorld H0: SkillOpt-Aligned Task Strategy

## Interface

Solve household tasks using text observations and the admissible commands
provided for the current step. Choose a command from that list, preserving
its exact object names and numbers. Return <think>...</think> followed by
<action>one command</action>. The environment alone determines task success.

## Six Task Families

| Task | Ordered plan |
| --- | --- |
| Pick and place | Locate the object, take it, reach the destination, place it. |
| Pick two and place | Locate, take and deliver the first object; repeat for a distinct second object. |
| Examine in light | Find and hold the object, reach the specified lamp, use the lamp. |
| Clean and place | Find and take the object, clean it at the sink, deliver it to the destination. |
| Heat and place | Find and take the object, heat it at the microwave, deliver it to the destination. |
| Cool and place | Find and take the object, cool it at the fridge, deliver it to the destination. |

## Decision Priorities

1. Read the goal and identify the target, required state change, destination
   and number of objects. Work through locate, acquire, transform and deliver.
2. Search surfaces and containers systematically. Open a closed container
   before concluding that it has no target. Keep a short frontier of
   unsearched locations from the latest evidence; once several checks have
   been negative or the frontier is exhausted, move to a genuinely new
   plausible location rather than revisiting unchanged spots.
3. When the required object is visible and reachable and your hand is free,
   take it immediately instead of continuing to wander.
4. Finish any required cleaning, heating or cooling before final delivery.
   Check the feedback; issuing a command does not prove its success.
5. Once carrying the required object in the required state, go directly to
   the destination and place it using the command in the admissible list.
6. Track distinct objects delivered and the remaining count. Do not confuse
   one successful placement with completion of a two-object task.
7. Avoid action loops. Do not issue the same command more than twice in a
   row without changed evidence. If a command produces "Nothing happens."
   or otherwise leaves the state unchanged, treat it as a failed
   transition and replan immediately by checking preconditions or searching
   an unexplored location rather than repeatedly toggling or examining an
   object.
8. Do not invent commands or object identifiers. A listed action is executable,
   not necessarily useful; choose it according to the task and current evidence.

## Common Errors

Do not ignore a reachable target, skip a required transformation, repeatedly
search an unchanged container, or declare completion without observed evidence.
Inventory and location may change after an action; use the latest feedback.
\end{harnesslisting}

\clearpage
\begin{harnesslisting}{ALFWorld (EnactOpt): Context Code}
HISTORY_LENGTH = 50


def _clip(text, limit=180):
    text = "" if text is None else str(text)
    text = text.replace("\n", " ")
    if len(text) > limit:
        return text[:limit].rstrip() + "..."
    return text


def build_context(task, history, observation, skill):
    recent = history[-HISTORY_LENGTH:]
    start = len(history) - len(recent)
    lines = []
    for index, record in enumerate(recent, start + 1):
        lines.append(
            f"[Observation {index}: '{_clip(record.get('observation', ''))}', "
            f"Action {index}: '{_clip(record.get('action', ''))}']"
        )
    return {
        "history_text": "\n".join(lines),
        "retained_turns": list(range(start, len(history))),
    }
\end{harnesslisting}

\clearpage
\subsection{EAI}
\label{app:eai-harness-artifacts}

\subsubsection{Evolved Harness}
\label{app:eai-evolved}
The following boxes reproduce the evolved EAI skill and
context-construction code.
% 中文对照：
% 以下文本框保留 EAI 进化后的技能与上下文构造代码。

\begin{harnesslisting}{EAI (EnactOpt): Skill}
Generate a nonempty ordered action plan from the current state using only supported action names and exact object IDs.
When feedback reports format_error or parse_error before any actions execute, repair the JSON envelope first; do not change the task logic until the submission is parseable.
If proximity to a target object is uncertain, WALK or FIND to the exact object before any direct interaction, and do not assume room membership implies range.
Treat facing, proximity, and possession as separate prerequisites: if LOOKAT/WATCH fails, TURNTO the same target first and then retry the observation before continuing.
When execution feedback reports a failure, use the updated state and specific failure cause to repair that unmet precondition instead of replaying the same blocked prefix; if that branch remains blocked, switch to another valid branch rather than repeating the blocker.
Prefer the shortest legal sequence that matches the visible affordances, and keep repeated action-goal actions in the required order.
Keep attempting while budget remains; do not stop after a review or failed correction, do not declare success or abandon the task, and let the environment's success signal end execution.
\end{harnesslisting}

\clearpage
\begin{harnesslisting}{EAI (EnactOpt): Context Code}
import json


def _render_value(value):
    if value is None:
        return 'null'
    if isinstance(value, str):
        text = value.strip()
        return text if text else '<empty>'
    try:
        return json.dumps(value, ensure_ascii=True, sort_keys=True, indent=2)
    except TypeError:
        return str(value)


def _section(title, value):
    return f'{title}:\n{_render_value(value)}'


def build_context(task, history, observation, skill):
    prompt = task['llm_prompt']
    if not isinstance(prompt, str) or not prompt.strip():
        raise ValueError('The complete official baseline prompt is required')
    if not isinstance(skill, str) or not skill.strip():
        raise ValueError('A non-empty Skill is required')

    latest_history = None
    if isinstance(history, list) and history:
        latest_history = history[-1]

    runtime_sections = [
        _section('Current observation (authoritative state)', observation),
    ]
    if latest_history is not None:
        runtime_sections.append(_section('Latest decision history item', latest_history))
    runtime_sections.append(_section('Full decision history', history))

    return {
        'messages': [
            {'role': 'user', 'content': 'Official task prompt (initial problem statement):\n' + prompt},
            {'role': 'user', 'content': 'Skill guidance:\n' + skill},
            {
                'role': 'user',
                'content': 'Runtime evidence (authoritative current state; only executed actions have effects; use the latest feedback and state, not the stale initial prompt):\n'
                + '\n\n'.join(runtime_sections),
            },
        ]
    }
\end{harnesslisting}

\end{document}